%% file: main_eccv.tex
\documentclass[runningheads]{llncs}

\usepackage[mobile]{eccv}

\usepackage{eccvabbrv}

\usepackage{graphicx}
\usepackage{booktabs}

\usepackage[accsupp]{axessibility}  

\input{misc/imports}
\input{misc/preamble}

\usepackage{hyperref}

\usepackage{orcidlink}

\begin{document}

\title{\paperTitle} 

\titlerunning{\paperTitleRunning}

\author{\paperAuthors{}}

\authorrunning{\runningAuthors{}}

\institute{\paperInstitute}

\maketitle

\input{sec/0_abstract}
\input{sec/1_intro}
\input{sec/2_related_works}
\input{sec/3_method}
\input{sec/4_experiments}
\input{sec/5_conclusion}

\section*{Acknowledgements}
We would like to thank Ryan Teehan, Vivian Lee and Kyunghyun Cho for the helpful discussions and advice on improving this manuscript. We thank NYU for providing some of the GPU resources to support this work.

%
%
\bibliographystyle{splncs04}
\bibliography{main}

\clearpage
\input{sec/supp/0_suppl}

\end{document}

%% file: misc/imports.tex
\usepackage[ruled,vlined,linesnumbered]{algorithm2e}
\usepackage{comment}
\usepackage{graphicx}
\usepackage{hhline}
\usepackage[table,dvipsnames]{xcolor}
\usepackage{bbm}
\usepackage{makecell}
\usepackage{amssymb}
\usepackage{pifont}
\usepackage{caption}
\usepackage{subcaption}
\usepackage{multirow}
\usepackage[pagebackref,breaklinks,colorlinks,allcolors=cvprblue]{hyperref}
\usepackage{tabularx}
\usepackage{wrapfig}
\usepackage{xparse}

%% file: misc/preamble.tex
\definecolor{cvprblue}{rgb}{0.21,0.49,0.74}

\newcommand{\Our}[0]{SN}
\newcommand{\Ours}[0]{SNs}

\newcommand{\oursFull}[0]{super neurons}

\newcommand{\OursFull}[0]{Super Neurons}
\newcommand{\savs}[0]{SAVs}

\newcommand{\savsFull}[0]{sparse attention vectors}

\newcommand{\SavsFull}[0]{Sparse Attention Vectors}

\newcommand{\mcqFull}[0]{multiple choice question}

\newcommand{\mcq}[0]{MCQ}

\newcommand{\eos}[0]{\texttt{<eos>}}
\newcommand{\vqa}[0]{VQA}
\newcommand{\bvqa}[0]{B-VQA}
\newcommand{\bvqaFull}[0]{binary visual question answer}
\newcommand{\cvqa}[0]{categorical VQA}

\newcommand{\cvqaFull}[0]{categorical visual question answer}

\newcommand{\sota}[0]{state of the art}

\newcommand{\llm}[0]{LLM}

\newcommand{\llmFull}[0]{large language model}

\newcommand{\vlm}[0]{VLM}
\newcommand{\vlms}[0]{VLMs}

\newcommand{\VlmFull}[0]{Vision-language model}
\newcommand{\vlmsFull}[0]{vision-language models}

\newcommand{\arFull}[0]{agreement rate}
\newcommand{\ArFull}[0]{Agreement rate}
\newcommand{\ar}[0]{AR}
\newcommand{\arAt}[1]{AR@#1}

\newcommand{\mathData}[0]{\mathcal{D}}
\newcommand{\mathTask}[0]{\Gamma}
\newcommand{\mathProbing}[0]{p}
\newcommand{\mathVal}[0]{v}
\newcommand{\mathNumSamples}[0]{N}
\newcommand{\mathSampleIdx}[0]{n}
\newcommand{\mathImgDim}[0]{H \times W \times 3}
\newcommand{\mathImgSeq}[0]{s_\text{vis}}
\newcommand{\mathTxtSeq}[0]{s_\text{txt}}
\newcommand{\mathLatentDim}[0]{D}
\newcommand{\mathNumLayers}[0]{L}
\newcommand{\mathLayerIdx}[0]{l}

\newcommand{\mathImg}[0]{\mathbf{I}}
\newcommand{\mathImgTaskProbingSample}[0]{\mathImg^{(\mathTask, \mathProbing)}_\mathSampleIdx}
\newcommand{\mathTxtTaskProbingSample}[0]{\mathTxt^{(\mathTask, \mathProbing)}_\mathSampleIdx}
\newcommand{\mathImgTaskValSample}[0]{\mathImg^{(\mathTask, \mathVal)}_\mathSampleIdx}
\newcommand{\mathTxtTaskValSample}[0]{\mathTxt^{(\mathTask, \mathVal)}_\mathSampleIdx}
\newcommand{\mathActTaskValSampleIdx}[0]{\mathAct^{(\mathTask, \mathVal)}_\mathSampleIdx}
\newcommand{\mathTxt}[0]{\mathbf{T}}
\newcommand{\mathGt}[0]{\mathbf{Y}}
\newcommand{\mathVLMOut}[0]{\mathbf{\hat{Y}}}
\newcommand{\mathAct}[0]{\mathbf{H}}
\newcommand{\mathActTaskProbing}[0]{\mathbf{H}^{(\mathTask, \mathProbing)}}
\newcommand{\mathActTaskProbingSampleIdx}[0]{\mathbf{H}^{(\mathTask, \mathProbing)}_\mathSampleIdx}
\newcommand{\mathActTaskVal}[0]{\mathbf{H}^{(\mathTask, \mathVal)}}
\newcommand{\mathMetric}[0]{\mu}
\newcommand{\snThresh}[0]{SN\(t\)}
\newcommand{\mathScore}[0]{\mathbf{S}}
\newcommand{\mathScoreTaskMetric}[0]{\mathScore^{(\mathTask, \mathProbing)}_\mathMetric}
\newcommand{\mathSn}[0]{\mathbf{\Sigma}}
\newcommand{\mathActThresh}[0]{\alpha}
\newcommand{\mathNumSn}[0]{K}
\newcommand{\mathSnIdx}[0]{k}
\newcommand{\mathSnPred}[0]{\mathbf{\hat{Y}}^\star}
\newcommand{\indexFn}{idx}
\newcommand{\llavaSmall}[0]{LLaVA-v1.5-7b}
\newcommand{\llavaSmallIt}[0]{\textit{LLaVA-v1.5-7b}}
\newcommand{\llavaLarge}[0]{LLaVA-v1.5-13b}

\newcommand{\qwenSmallIt}[0]{\textit{Qwen3-VL-4b-Instruct}}
\newcommand{\qwenSmall}[0]{Qwen3-VL-4b-Instruct}

\newcommand{\qwenLarge}[0]{Qwen3-VL-32b-Instruct}

\newcommand{\mathLLM}[0]{f}
\newcommand{\mathImgEnc}[0]{E_\text{vis}}
\newcommand{\mathTxtEnc}[0]{E_\text{txt}}

\newcommand{\mathProbingSet}[0]{\mathData^{(\mathTask, \mathProbing)}}
\newcommand{\mathValSet}[0]{\mathData^{(\mathTask, \mathVal)}}

\definecolor{lightGray}{gray}{0.9}
\definecolor{lightYellow}{HTML}{FFF2CC}

\newcommand{\benchIncrease}[1]{\scriptsize{\textcolor{ForestGreen}{#1}}}
\newcommand{\benchIncreaseFmt}[1]{%
  \makebox[0pt][l]{\textcolor{ForestGreen}{\tiny (+#1)}}%
}

\newcommand{\benchDecreaseFmt}[1]{%
  \makebox[0pt][l]{\textcolor{purple}{\tiny (-#1)}}%
}
\renewcommand{\benchIncreaseFmt}[1]{\textcolor{ForestGreen}{\tiny(+#1)}}
\renewcommand{\benchDecreaseFmt}[1]{\textcolor{purple}{\tiny(-#1)}}

\newcommand{\cmark}{\ding{51}}%
\newcommand{\xmark}{\ding{55}}%

\newcolumntype{g}{>{\columncolor{lightGray}}c}
\newcolumntype{G}{>{\columncolor{lightGray}}l}
\newcolumntype{y}{>{\columncolor{lightYellow}}c}
\newcolumntype{Y}{>{\columncolor{lightYellow}}l}

\newcommand{\paperTitle}[0]{Taking Shortcuts for Categorical VQA Using Super Neurons}

\newcommand{\paperAuthors}[0]{
Pierre Musacchio\inst{1}\orcidlink{0009-0008-4107-175X}
\and
Jaeyi Jeong\inst{2}\orcidlink{0009-0003-8106-8129}
\and
Dahun Kim\inst{3}\orcidlink{0000-0003-1776-6195}
\and Jaesik Park\inst{1}\orcidlink{0000-0001-5541-409X}
}
\newcommand{\runningAuthors}[0]{P.~Musacchio et al.}
\newcommand{\paperInstitute}[0]{Seoul National University \and
EPFL \and
Google Deepmind\\
\email{\{pmusacchio, jaesik.park\}@snu.ac.kr}
}
\newcommand{\paperTitleRunning}[0]{Super Neurons}

\newcommand{\StartDynFig}[3]{
  \ifdefined\maketitlesupplementary
    \begin{figure}[#1]
  \else
    \begin{wrapfigure}{#2}{#3}
  \fi
}
\newcommand{\EndDynFig}{
  \ifdefined\maketitlesupplementary
    \end{figure}
  \else
    \end{wrapfigure}
  \fi
}

\newcommand{\StartDynTab}[3]{
  \ifdefined\maketitlesupplementary
    \begin{table}[#1]
  \else
    \begin{wraptable}{#2}{#3}
  \fi
}
\newcommand{\EndDynTab}{
  \ifdefined\maketitlesupplementary
    \end{table}
  \else
    \end{wraptable}
  \fi
}

%% file: sec/0_abstract.tex
\begin{abstract}
    \SavsFull{} (\savs{}) have emerged as an excellent training-free alternative to supervised finetuning or low-rank adaptation to improve the performance of \vlmsFull{} (\vlms{}).
    At their heart, \savs{} select a few accurate attention heads for a task of interest and use them as classifiers, rather than relying on the model's prediction.
    In a similar spirit, we find that directly probing the raw activations of the \vlm{}, in the form of scalar values, is sufficient to yield accurate classifiers on diverse visually grounded downstream tasks.
    Shifting focus from attention vectors to scalar activations dramatically increases the search space for accurate parameters, allowing us to find more discriminative neurons immediately from the first generated token. We call such activations \OursFull{} (\Ours{}). In this probing setting, we discover that enough \Ours{} appear in the shallower layers of the \llmFull{} to allow for extreme early exiting from the first layer of the model at the first generated token. Compared to the original network, \Ours{} robustly improve the classification performance while achieving a speedup of up to 5.10\(\times\).
\end{abstract}

%% file: sec/1_intro.tex
\section{Introduction}
\label{sec:intro}

\input{figs/teaser}

\VlmFull{} (\vlm{}) are frontier models extending the generative capabilities of \llmFull{} (\llm{}) via visual grounding~\cite{liu23llava, liu2025nvila, bai2025qwen3vl, cheng24spatialrgpt}. Usually consisting of billions of parameters, these models retain extensive knowledge from internet-scale pretraining~\cite{brown20gpt3, touvron23llama2, dubey2024llama3, openai2023chatgpt, radford21clip}. Although remarkably effective, their complexity hinders attempts to understand how they operate at their core.

Current research on \vlm{} explainability and efficiency improvement mainly focuses on what could be called \emph{macro}-level representations. That is, multidimensional representations that are learned through aggregating information from interactions of the tokens in the model. The most famous examples lie in linear probing~\cite{skean25layerbylayer, yu25multimodalllmimagetasks} or attention map extraction~\cite{kang25fewheads, mitra25savs}. However, thanks to the over-parameterization of current \sota{} networks, we hypothesize that models accumulate such a tremendous amount of information over training that their individual activation scalars are sufficient to provide accurate answers to specific questions. We term these \textit{micro}-level representations.

Thus, we repurpose the neuron activations of the model into predictions via a simple training-free strategy inspired by~\cite{mitra25savs}. Analogously, we gather a probing dataset and perform an end-to-end \vlm{} inference on it. During the process, we store activations from the \llmFull{} (\llm{}) of the \vlm{}. However, instead of clustering attention heads, we directly convert the raw activations into classification predictions by thresholding them. We observe that this simple conversion scheme is enough for a subset of neurons to achieve high scores on conventional \cvqaFull{} metrics for a wide diversity of datasets. We subsequently deem them \emph{\OursFull{}} (\Ours{}). Surprisingly, \Ours{} obtain even better performance than the models themselves on a diverse suite of unseen \cvqa{} validation benchmarks. Since there are more raw activations than attention heads in the network (cf.~\cref{tab:search_space}), there are more chances to find \Ours{} that have desirable properties, such as better performance and robustness. Specifically, we discover that some \Ours{} located in shallower layers of the model preserve great performance even while the first token is being generated. This allows us to perform \emph{extreme early exit} \ie interrupt inference on the first layer of the \llm{} during the generation of the first token.

Our contributions are summarized as follows:
\begin{itemize}
    \item We shift the analysis from \emph{macro}-level representations (akin to attention vectors) to \emph{micro}-level ones (scalar activations). By doing so, we present a training-free approach that identifies high-scoring neurons in the \llm{} of the \vlms{},
    \item We comprehensively benchmark the probed neurons and find that they can serve as strong categorical classifiers, \emph{outperforming the base models themselves} on a diverse suite of \vqa{} benchmarks. We therefore call them \emph{\OursFull{}},
    \item We thoroughly investigate \Ours{} (discriminative power, location in the model, quantity, robustness) and introduce the \emph{\arFull{}} metric that quantifies the divergence between \Ours{} predictions and model predictions,
    \item As a byproduct, \Ours{} enable extreme early exit at inference time, providing a speedup of up to \(5.10\times\) while maintaining model-level performance.
\end{itemize}

%% file: figs/teaser.tex
\StartDynFig{t}{R}{0.6\linewidth}
    \centering
    \includegraphics[width=\linewidth]{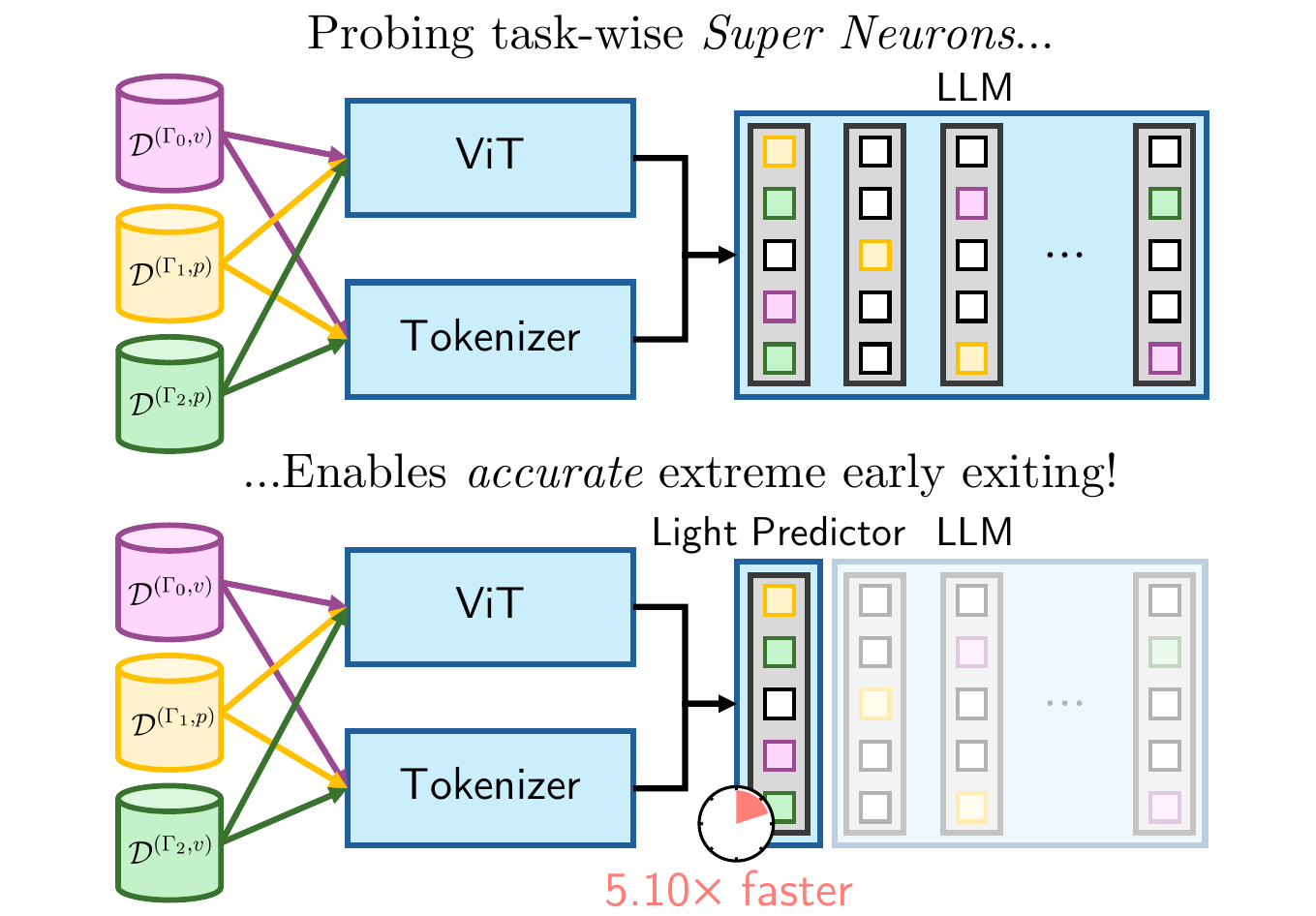}
    \caption{\textbf{Overview of our approach.}
    Our training-free scheme uncovers \OursFull{} (\Ours{}) via probing data.
    They robustly outperform the base model on a variety of \cvqa{} datasets.
    As a byproduct, they enable \emph{extreme early exiting} at the first layer of the \llm{} on the first generated token.
    Colored boxes in the \llm{} represent \Ours{} for their data types.}
    \label{fig:teaser}
\EndDynFig{}

%% file: sec/2_related_works.tex
\section{Related work}
\label{sec:related_works}

\paragraph{Efficient \vlms{}.}
A conventional approach to turn large \vlms{} to efficient models is to prune them at the parameter level, either by distillation~\cite{wang23efficientvlm} or by training a policy to search which weights to remove~\cite{liang2025efficientllava}. Pruning can also occur at the token level, usually via token similarity approaches~\cite{ye2025atpllava, jeddi2025yourvlmfaster, zhang2024fastervlm, cao2023pumer}, by estimating visual contribution~\cite{liu2025meteor}, or using scale-down approaches~\cite{liu2025nvila}. If the final objective is to improve performance, training a robust visual encoder is also a viable solution~\cite{tang25tulip, fu2023blink}. Some approaches diverge by considering early exit from the model, either in a supervised setting~\cite{bajpai2025free} or in a training-free manner by estimating layer-wise similarities~\cite{tang2023similarityexit}.
Recently, task vectors~\cite{hojel2024visualtaskvectors} have been leveraged in \vlms{} in the form of \savsFull{} to enable training-free improvement of \vlms{} in classification tasks~\cite{mitra25savs}.
Single modality convnets and LSTMs can rely on some of their weights for accurate prediction~\cite{le2012features, radford2017bytelstm}, but this remains to be shown for transformers, specifically when processing multimodal tokens.

\input{figs/ours_vs_savs}

While inspired by~\cite{mitra25savs}, we elect neural activations rather than clustering attention heads, shifting the representation of interest from a macro- to a micro-level (cf.~\cref{fig:ours_vs_savs}). This shift continues to hold properties reported in~\cite{le2012features, radford2017bytelstm, mitra25savs} (\eg better performance than the model itself), while being robust to prompt variations and distribution shifts, enabling inference on diverse \vqa{} tasks and extreme early stopping.

Although we solely focus on \cvqa{} tasks, we propose a training-free approach that identifies expert neural activations and establishes a set of \Ours{} that solves the task accurately and robustly, without relying on token similarity or altering model weights. After discovery, we substantially improve the runtime performance of \vlm{} using extreme early exit, as early as the first layer.

\paragraph{Explainable \vlms{}.} Model explainability is a fundamental challenge for the deployment of \vlms{} in the real world. Since \vlms{} are increasingly being adopted as master operators in robotics~\cite{kim2024openvla, black2410pi0, zitkovich2023rt2}, guardrails must be set up to ensure the security of their behavior. Substantial work has led to a better understanding of how attention, which \vlms{} are usually built on, behaves. Notably, CLIP-Dissects proposes tagging each neuron in the transformer with a concept~\cite{oikarinen2022clipdissect}, showing that transformers learn more complex patterns as the representation is forwarded down its layers. Efforts have also been directed towards understanding attention sinks~\cite{oquadb24dinov2, darcet23registers, kang2025see}. Moreover, due to the extensive number of attention operations in the \llm{} of the \vlm{}, the community has reported the emergence of object-aligned attention maps in the transformer decoder of the architecture~\cite{kang25fewheads}. Linear probing approaches tend to show that the \vlm{} generates its answer based on different stages of reasoning~\cite{yu25multimodalllmimagetasks}, yet, these stages do not seem to be monolithic~\cite{skean25layerbylayer, mitra25savs}. Sparse autoencoders has shown that some specific neurons hold object-specific concepts~\cite{huben24sae, templeton24goldenbridge}.

In our work, we propose studying the capabilities of individual neurons without adding a single learning component that could alter the understanding of their function. By repurposing raw activations as categorical predictions, we show that \vlms{} possess expert neurons across a diverse set of tasks. Analyzing the location of the emergence of these neurons helps us better understand that the LLM is in principal capable of answering a question sometimes as early as in the \emph{first layer} of the \llm{} when generating the \emph{first token} of the answer. We also investigate to what extent \Ours{} and the model disagree by introducing the \arFull{} (\ar{}) metric. Robustness experiments suggest that \Ours{} are \emph{not} exploiting spurious correlations of the input data and generalize to new distributions or neighboring prompts, suggesting the universality of our approach.

%% file: figs/ours_vs_savs.tex
\StartDynFig{t}{r}{0.6\linewidth}
        \centering
        \begin{subfigure}[t]{0.9\linewidth}
            \centering
            \caption{Architecture comparison between SAVs and SNs.}
            \label{fig:ours_vs_savs}
            \includegraphics[width=\linewidth]{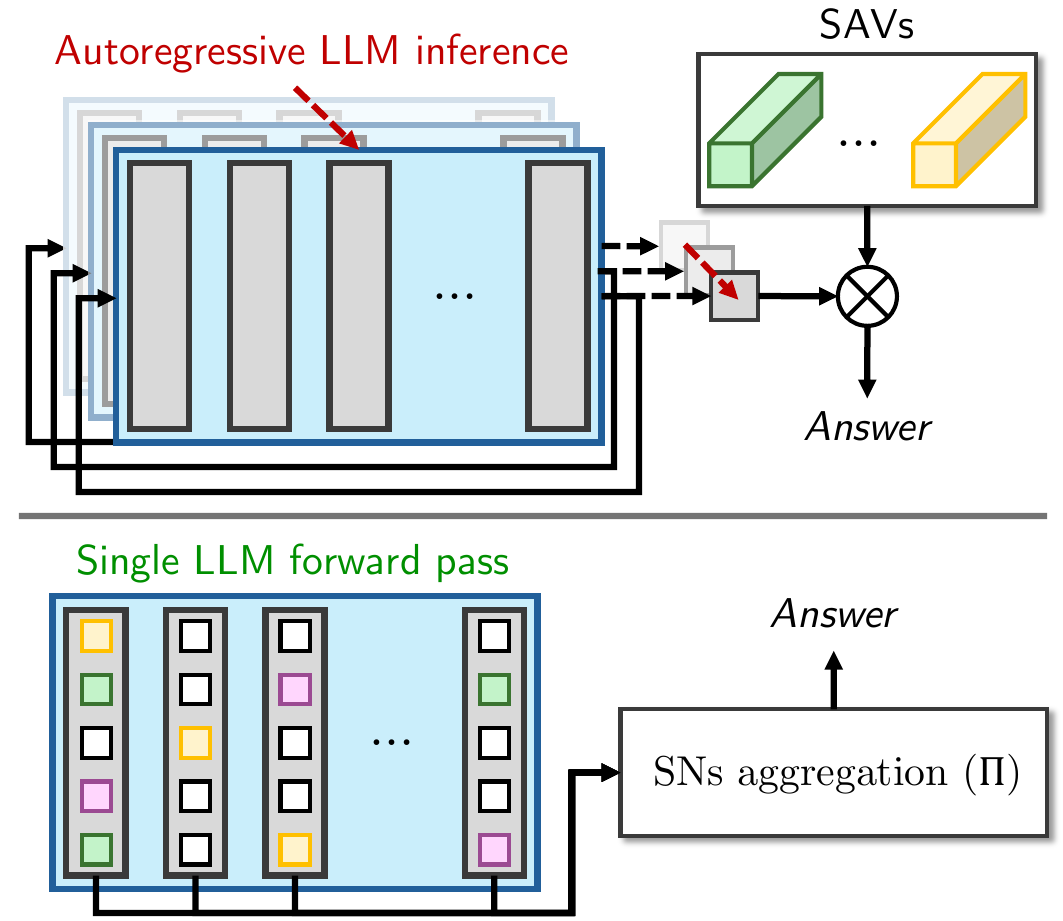}
        \end{subfigure}
        
        \begin{subfigure}[t]{\linewidth}
        \centering
        \caption{Search space comparison.}
        \label{tab:search_space}
        \resizebox{\linewidth}{!}{
        \begin{tabular}{lll}
             \toprule
             Method & Search target & Search space size \\
             \midrule
             \midrule
             SAVs~\cite{mitra25savs} & Attention vectors & Layers \(\times\) Heads \(=32 \times 32 =1024\) \\
             \rowcolor{lightYellow}\Ours{} (Ours) & Activation scalars & Layers \(\times\) Dim. \(= 32 \times 4096 = \mathbf{131,072}\) \\
             \bottomrule
        \end{tabular}
        }
        \end{subfigure}
        \caption{\textbf{Comparison between SAVs and ours.} We show architectural divergences in~\cref{fig:ours_vs_savs} and resulting search spaces of the two approaches for \llavaSmall{} in~\cref{tab:search_space}.}
\EndDynFig{}

%% file: sec/3_method.tex
\input{algo/find_sn}
\section{Method}
\label{sec:method}

\subsection{Preliminaries}
\label{sec:preliminaries}
\paragraph{Notations.} We define a VLM as the combination of vision and text encoders \(\mathImgEnc : \mathbb{R}^{\mathImgDim} \rightarrow \mathbb{R}^{\mathImgSeq \times \mathLatentDim}\) and \(\mathTxtEnc : \mathbb{N}^s \rightarrow \mathbb{R}^{\mathTxtSeq \times \mathLatentDim}\) that feed their output to an LLM \(\mathLLM : \mathbb{R}^{(\mathImgSeq + \mathTxtSeq + 1) \times \mathLatentDim} \rightarrow \mathbb{R}^{(\mathImgSeq + \mathTxtSeq + 1) \times \mathLatentDim}\). 
Given a grounding image \(\mathImg \in \mathbb{R}^{\mathImgDim}\) and a text prompt \(\mathTxt \in \mathbb{N}^s\), a VLM forward pass is defined as follows:
\begin{equation}
    \mathVLMOut^{s + 1} = \mathLLM (\mathImgEnc(\mathImg); \mathTxtEnc(\mathTxt)).
\end{equation}
Here, \(\mathVLMOut^{s + 1}\) can be auto-regressively fed into \(\mathLLM\). This process ends when the LLM generates an \eos{} token.
Moreover, given an \(\mathNumLayers \in \mathbb{N}\) layered LLM, we denote \(\mathAct_\mathLayerIdx \in \mathbb{R}^\mathLatentDim\) the activation extracted from the \(\mathLayerIdx\)-th layer. For clarity, we omit the subscript when referring to the full set of \(\mathNumLayers\) activations, \ie \(\mathAct \in \mathbb{R}^{\mathNumLayers \times \mathLatentDim}\).

\input{algo/sn_inference}
\paragraph{Problem.}
Conventional VLMs are built from an LLM architecture that contains billions of parameters, processing both vision and text tokens. We hypothesize that this parametric scale is reasonable for individual neurons to hold critical information about the answer for a given text-image pair and that we do not necessarily need the full model to answer the question. Specifically, we claim that \textit{some scalar activations are sufficient to provide a satisfactory answer, on par or even better than the full model itself}. We call these hypothetical neuron outputs \textit{\OursFull{}} (\Ours{}). Inspired by~\cite{mitra25savs}, our work provides a simple setup to discover such \Ours{} for categorical \vqa{}.

Uncovering \Ours{} can be thought of as a three-step process. First, we gather a \textit{probing set}. We then perform a forward pass of the network on the probing set to uncover neurons that have high activations on it based on a metric to optimize. 
Finally, we evaluate \Ours{} on the validation set of the dataset to assess their performance. We provide the complete algorithm used to discover \Ours{} in \cref{alg:find_superneurons}.

\subsection{\OursFull{}}
\label{sec:\OursFull{}}
\paragraph{Probing set.} Formally, we identify a task \(\mathTask\) to solve and gather a probing dataset \(\mathProbingSet = \left\{ \mathProbingSet_\mathSampleIdx \right \}_{\mathSampleIdx = 0}^{\mathNumSamples \in \mathbb{N}}\). Here, \(\mathProbing\) stands for \textit{probing set}. This probing set is typically built from training data used to optimize a model for \(\mathTask\). We gather the full model activations for each of the vision-text pairs of the probing set \(\left( \mathImg^{(\mathTask, \mathProbing)}_\mathSampleIdx, \mathTxt^{(\mathTask, \mathProbing)}_\mathSampleIdx \right) \in \mathProbingSet_\mathSampleIdx\):
\begin{equation}
    \mathAct^{(\mathTask, \mathProbing)}_\mathSampleIdx = \mathLLM (\mathImgEnc (\mathImg^{(\mathTask, \mathProbing)}_\mathSampleIdx); \mathTxtEnc (\mathTxt^{(\mathTask, \mathProbing)}_\mathSampleIdx)) \in \mathbb{R}^{\mathNumLayers \times \mathLatentDim}.
\end{equation}
Taking into account all samples in the dataset, we note \(\mathActTaskProbing \in \mathbb{R}^{\mathNumSamples \times \mathNumLayers \times \mathLatentDim}\) the tensor of all activations.

\paragraph{Discovering \OursFull{}.}
The key idea is to directly convert the raw activations into binary predictions. Hence, we introduce a threshold variable \(\mathActThresh \in \mathbb{R}\) responsible for binarizing the raw activations:
\begin{equation}
    \label{eq:act_thresh}
    \mathActTaskProbing > \mathActThresh.
\end{equation}
We detail how we instantiate this value in~\cref{sec:experimental_setting}.

We proceed to evaluate each neuron on the full probing set using a predetermined metric \(\mathMetric\) to acquire neuron-level statistics:
\begin{equation}
    \mathScoreTaskMetric = \mathMetric \left( \mathAct^{(\mathTask, \mathProbing)}_\mathSampleIdx, \mathGt_\mathSampleIdx \right), \quad \forall \mathSampleIdx \in \{0, \dots, \mathNumSamples \},
\end{equation}
where \(\mathGt_\mathSampleIdx\) is the ground-truth for the \(\mathSampleIdx\)-th data sample and \(\mathScoreTaskMetric \in [0, 1]^{\mathNumSamples \times \mathNumLayers}\) represents the neuron-wise scores for task \(\mathTask\) on the probing set \(\mathProbing\) with respect to a metric \(\mathMetric\). Conventional metrics are usually normalized from 0 to 1. Therefore, we consider this formulation.

At this stage, we identify neurons as \oursFull{} if they fall above a specific predetermined metric threshold. We call this threshold \(\text{\snThresh{}} \in [0, 1]\). Thus, the final \Ours{} are selected as follows:
\begin{equation}
    \mathSn = \text{\indexFn{}} \mathAct^{(\mathTask, \mathProbing)}[\mathScoreTaskMetric > \text{\snThresh{}}].
\end{equation}
Hence, \(\mathSn\) represents the index map of the thresholded \Ours{} for a given \snThresh{} and \indexFn{} is a function that returns the indices of the tensor values that meet the thresholding requirement.

\input{figs/act_thresh_range}

\paragraph{Evaluating \OursFull{} on validation data.}
\input{figs/empirical_analysis}
Once \(\mathSn\) is obtained, we perform the inference on the validation set of \(\mathTask\) denoted \(\mathValSet = \left \{ \mathValSet_\mathSampleIdx \right \}_{\mathSampleIdx = 0}^\mathNumSamples\), where \(\mathVal\) stands for \textit{validation set}. As with the probing set, we extract \(\mathActTaskVal\) and only select \Our{} activations from indexing on \(\mathSn\):
\begin{equation}
    \mathSnPred = \mathActTaskVal[\mathSn] > \mathActThresh,
\end{equation}
where \(\mathSnPred \in \{0, 1\}^{\mathNumSn}\) and \(\mathNumSn \in \mathbb{N}\) denotes the number of selected \Ours{}. Since \(\mathNumSn\) is usually larger than 1, we finally aggregate all the \Our{} predictions into a single final prediction using an aggregation function \(\mathbf{\Pi}\). For this, we use two different strategies. Either, we simply average all \Our{} predictions or we perform majority voting. We provide the inference routine in~\cref{alg:sn_inference}.

\subsection{\ArFull{}}
\label{sec:agreement_rate}
To measure how much \Ours{} diverge from the predictions of the model, we introduce the \arFull{} (\ar{}) metric. Conceptually, \ar{} aims at quantifying the frequency at which \Ours{} and the model have the same answers.
We define \ar{} as follows:
\begin{equation}
    \text{\ar{}} = \frac{1}{\mathNumSamples \mathNumSn} \sum_{\mathSampleIdx = 0}^{\mathNumSamples} \sum_{\mathSnIdx =  0}^\mathNumSn \mathbbm{1} \left( \mathSnPred_{\mathSampleIdx, \mathSnIdx} = \mathVLMOut_\mathSampleIdx \right) \in [0, 1].
\end{equation}
Here, \(\mathbbm{1}\) is the indicator function and we subscript the \Ours{} index by \(\mathSnIdx\), while subscripting the data samples by \(\mathSampleIdx\). Note that \ar{} can be obtained for different \snThresh{} thresholds. Thus, we also denote the metric with a suffix specifying the \snThresh{} \eg if \(\text{\snThresh{}}=0.8\), then \arAt{0.8} is the agreement rate across all \Ours{} whose accuracy exceeds 0.8 on the set of interest.

%% file: algo/find_sn.tex
\begin{algorithm}[t]
\caption{\textsc{\OursFull{} Extraction}}
\label{alg:find_superneurons}
\KwData{\\
\(\mathProbingSet\), probing dataset for task \(\mathTask\);\\
\(\mathImgEnc\) and \(\mathTxtEnc\), vision and text encoders of the VLM; \\ 
\(\mathLLM\), LLM of the VLM; \\
\(\mathActThresh\), activation threshold parameter; \\
\(\mathMetric\), a function that computes a metric; \\
\(\mathGt\), ground-truths; \\
\snThresh{}, \oursFull{} threshold.
}

\(\mathActTaskProbing \gets \{\}\)

\For{\( \mathImgTaskProbingSample, \mathTxtTaskProbingSample \in \mathProbingSet\)}{
    \(\mathActTaskProbingSampleIdx \gets \mathLLM (\mathImgEnc (\mathImg^{(\mathTask, \mathProbing)}_\mathSampleIdx); \mathTxtEnc (\mathTxt^{(\mathTask, \mathProbing)}_\mathSampleIdx))\) \;
    \(\mathActTaskProbing\).\textsc{Append}\((\mathActTaskProbingSampleIdx)\) \;
}
\( \mathActTaskProbing \gets \mathActTaskProbing > \mathActThresh{} \) \;
\( \mathScoreTaskMetric \gets \mathMetric \left( \mathActTaskProbing, \mathGt \right) \) \;
\( \mathSn = \text{\indexFn{}} \mathAct^{(\mathTask, \mathProbing)}[\mathScoreTaskMetric > \text{\snThresh{}}] \) \;
\Return \(\mathSn\) \;
\end{algorithm}

%% file: algo/sn_inference.tex
\begin{algorithm}[t]
\caption{\textsc{\OursFull{} Inference}}
\label{alg:sn_inference}
\KwData{\\
\(\mathImgEnc\) and \(\mathTxtEnc\), vision and text encoders of the VLM; \\ 
\(\mathLLM\), LLM of the VLM; \\
\(\mathActThresh\), activation threshold parameter; \\
\snThresh{}, \oursFull{} threshold; \\
\(\mathSn\), indexing of previously probed \Ours{}; \\
\(\mathbf{\Pi}\), an aggregation function.
}

\(\mathActTaskVal \gets \{\}\)

\For{\( \mathImgTaskValSample, \mathTxtTaskValSample \in \mathValSet\)}{
    \(\mathActTaskValSampleIdx \gets \mathLLM (\mathImgEnc (\mathImg^{(\mathTask, \mathVal)}_\mathSampleIdx); \mathTxtEnc (\mathTxt^{(\mathTask, \mathVal)}_\mathSampleIdx))\) \;
    \(\mathActTaskVal\).\textsc{Append}\((\mathActTaskProbingSampleIdx)\) \;
}
\( \mathActTaskVal \gets \mathActTaskVal > \mathActThresh \) \;
\Return \( \mathbf{\Pi}(\mathActTaskVal[\mathSn])\) \;
\end{algorithm}

%% file: figs/act_thresh_range.tex
\ifdefined\maketitlesupplementary
    \StartDynFig{t}{R}{0.5\linewidth}
        \centering
        \includegraphics[width=\linewidth]{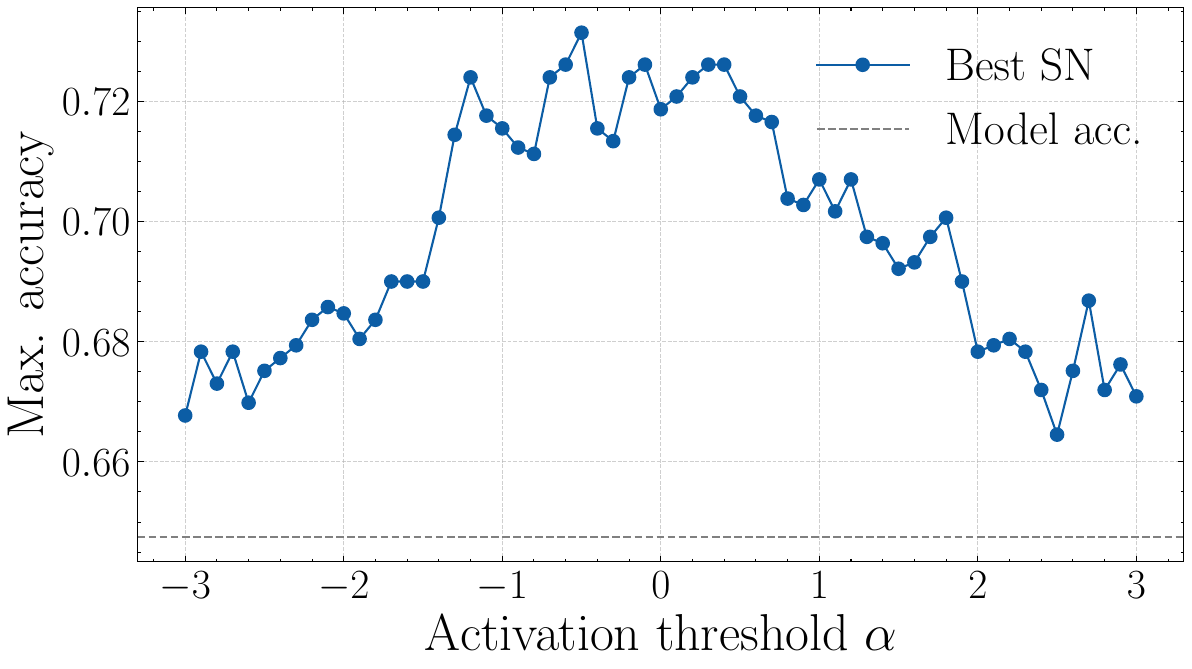}
        \caption{\textbf{Maximum accuracy obtained on the \textsc{VizWiz} probing set with respect to different \(\mathActThresh\).} We evaluate \(\mathActThresh\) over the range \(\mathActThresh \in [-3, 3]\) with a step of \(0.1\) and report the single best performing found \Our{}. The maximum accuracy peaks around \(\mathActThresh = 0\).}
        \label{fig:act_thresh_range}
    \EndDynFig{}
\fi

%% file: figs/empirical_analysis.tex
\ifdefined\maketitlesupplementary
\else
    \begin{figure}[t]
        \centering
        \begin{subfigure}[t]{0.5\linewidth}
        \includegraphics[width=\linewidth]{assets/vizwiz/max_accuracy_per_threshold.pdf}
        \caption{Accuracy w.r.t. the activation threshold \(\mathActThresh\).}
        \label{fig:act_thresh_range}
        \end{subfigure}%
        ~
        \begin{subfigure}[t]{0.5\linewidth}
        \includegraphics[width=\linewidth]{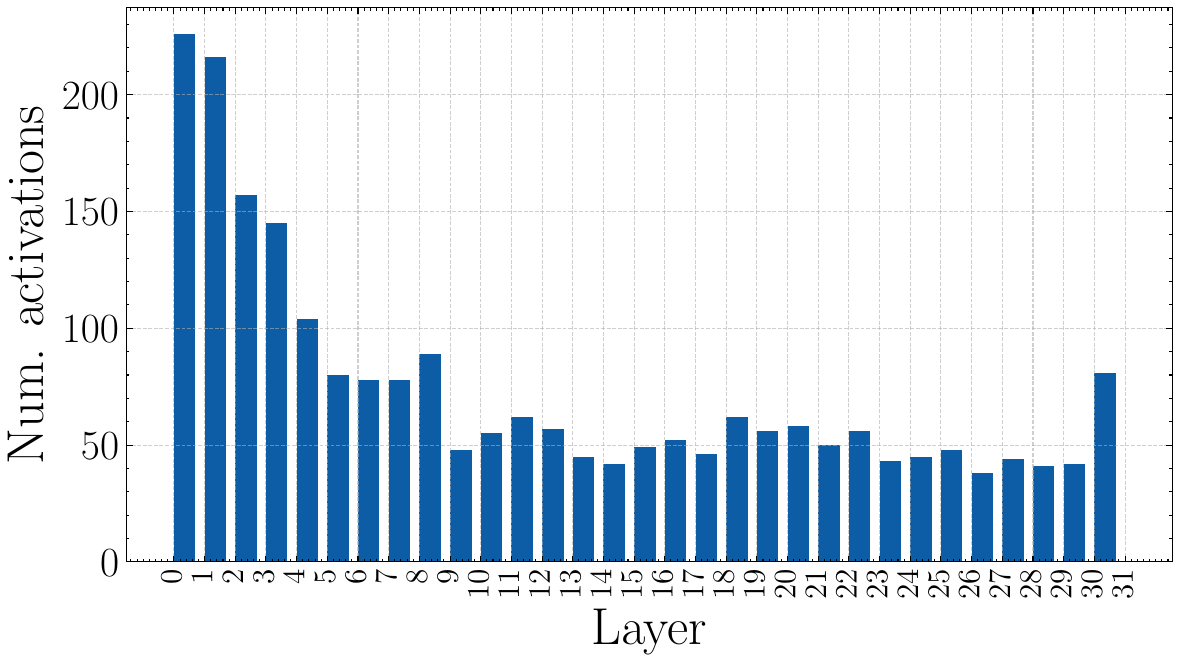}
        \caption{Number of \Ours{} per layer.}
        \label{fig:neurons_per_layer}
        \end{subfigure}
        \caption{\textbf{Empirical analysis of \Ours{}.} \Cref{fig:act_thresh_range} shows the maximum accuracy on the probing set with respect to different \(\mathActThresh\). We evaluate \(\mathActThresh\) over the range \(\mathActThresh \in [-3, 3]\). The maximum accuracy peaks around \(\mathActThresh = 0\). \Cref{fig:neurons_per_layer} records the number of found \Ours{} that obtain a better accuracy than the model in each layer. We use \llavaSmall{} on \textsc{VizWiz} for both figures.}
        \label{fig:empirical_analysis}
    \end{figure}
\fi

%% file: sec/4_experiments.tex
\section{Experiments}
\label{sec:experiments}

\input{figs/sn_per_layers}
\input{tables/probing_results}

\subsection{Datasets}
We validate our approach on seven diverse categorical \vqa{} datasets:
\begin{itemize}
    \item \textsc{Pope}, for object hallucination~\cite{li2023pope},
    \item \textsc{InstaOrder (Occ.)}, for occlusion understanding~\cite{lee22instaorder, musacchio2025instaformer},
    \item \textsc{InstaOrder (Depth)}, for depth understanding,
    \item \textsc{VizWiz}, for broad visual understanding~\cite{gurari2018vizwiz},
    \item \textsc{Clevr}, a synthetic dataset for geometrical understanding~\cite{johnson2017clevr},
    \item \textsc{A-OKVQA}, a general knowledge \mcqFull{} (\mcq{}) dataset~\cite{schwenk2022aokvqa},
    \item \textsc{ScienceQa}, an \mcq{} dataset for mathematics and scientific reasoning~\cite{lu22scienceqa}.
\end{itemize}
We provide details of each dataset, along with their respective prompt templates, in Appendix~\ref{sup:datasets_and_prompts}.

\subsection{Evaluation protocol}

\paragraph{Probing set.}
\input{tables/val_results}
\input{figs/agreement_rate}
We construct a probing set of size \(\mathNumSamples = 3,000\) for all datasets, randomly sampled from their respective training sets. By probing samples from the training set, we ensure there is \emph{no overlap} between the probing and validation data. We make a single exception for \textsc{VizWiz}, which contains fewer than 1K categorical \vqa{} in its train set.
We convert A-OKVQA and \textsc{ScienceQa} as a series of binary questions allowing our method to be applied indistinguishably as for other datasets. We balance the probing set of each dataset to ensure that each categorical class is evenly represented to avoid biasing the selected \Ours{}.

\paragraph{Models.}
\input{tables/sns_vs_savs}
We evaluate two well-established models in the \vlm{} landscape to emphasize the universal plug-and-play nature of our approach. We choose \llavaSmall{} since it is a cornerstone \vlm{} and has been widely adopted and modified~\cite{liu23llava, liu2025nvila}.
We also experiment on the more recently released \qwenSmall{} as its capabilities are known to be better than \llavaSmall{} while being significantly smaller~\cite{bai2025qwen3vl}. Finally, we conduct scaling-up experiments using \llavaLarge{} and \qwenLarge{}. Unless specified otherwise, we use the default model configuration in all cases. Further information can be found in appendix~\ref{sup:baseline_comparison}.

\paragraph{Experimental setting.}
\label{sec:experimental_setting}
Unless mentioned otherwise, we use NVIDIA RTX A6000 GPUs for our experiments.
The extraction of \Ours{} is training-free and simply requires the collection of raw activations from the model. Split across 8 GPUs, this only requires about 4 minutes of runtime for \llavaSmall{}.
To find the optimal \(\mathActThresh\) activation threshold, we first compute the mean activation across 3K randomly sampled \vqa{} in the \textsc{Pope}-style format, yielding 0.0083.
We also empirically pick different \(\mathActThresh\) values in the probing set of \textsc{VizWiz} in~\cref{fig:act_thresh_range}.
Interestingly, all tested values provide accuracy above the model itself.
Nevertheless, this figure confirms that the maximum accuracy peaks around \(\mathActThresh = 0\). Thus, we use \(\mathActThresh = 0\) for all experiments.
After running~\cref{alg:find_superneurons} on the probing set of each dataset, we choose the appropriate \snThresh{} by sweeping across values that are up to 3 points lower using a step size of 1 for \llavaSmall{} and a step of 0.1 for \qwenSmall{}. We use 128 max. generated tokens, set temperature to 0, use 1 beam, and don't leverage stop strings in all experiments, unless stated otherwise.
We detail the configurations of the models in appendix~\ref{sup:baseline_comparison}.

\input{tables/scale_up}
\paragraph{Metrics.}
To account for the fact that \vqa{} benchmarks can be imbalanced, not only do we report accuracy, as previous works do, but also compute precision, recall, and F1 score. This allows us to better estimate the predictive capabilities of all benchmarked methods. We use a rule-based evaluation strategy for accurate and interpretable results. Accuracy, precision, recall and F1 are defined as follows:

\begin{align}
    \text{Accuracy} &= \frac{1}{\mathNumSamples} \sum_\mathSampleIdx \mathbbm{1} \left( \mathSnPred_\mathSampleIdx = \mathGt_\mathSampleIdx \right), \\
    \text{Precision} &= \frac{ \sum_\mathSampleIdx \mathbbm{1} \left( \mathSnPred_\mathSampleIdx = 1 \land \mathGt_\mathSampleIdx = 1 \right)}{\sum_\mathSampleIdx \mathbbm{1} \left( \mathSnPred_\mathSampleIdx = 1 \right)}, \\
    \text{Recall} &= \frac{ \sum_\mathSampleIdx \mathbbm{1} \left( \mathSnPred_\mathSampleIdx = 1 \land \mathGt_\mathSampleIdx = 1 \right)}{\sum_\mathSampleIdx \mathbbm{1} \left( \mathGt_\mathSampleIdx = 1 \right)}, \\
    \text{F1} &= \frac{2 \times \text{Precision} \times \text{Recall}}{\text{Precision} + \text{Recall}}.
\end{align}

Following the previously established notations, \(\mathNumSamples\) denotes the size of the dataset, \(\mathSampleIdx\) a sample index, \(\mathSnPred\) a prediction made from an \Our{} and \(\mathGt\) the ground-truth label. At probing time, these metrics can serve as a choice of \(\mathMetric\). We also compute them on the model output. During validation, we recompute these metrics on the model and the elected \Ours{} to obtain the numbers reported in the benchmarks.

\input{figs/perf_wrt_num_samples}

\subsection{Results}
\paragraph{\Ours{} extraction.}
\input{tables/runtime}
We start by gathering \Ours{} from the probing set. We report the performance of the \emph{single best} scoring neuron in~\cref{tab:probing_results}. In this table, we optimize for two main metrics \(\mathMetric = \{\text{accuracy, F1}\}\) and report their respective results in different rows.
We observe that the highest-scoring neuron surpasses the model on all probed datasets. We specifically note large differences on \textsc{VizWiz} and \textsc{InstaOrder (Occ.)}, tasks that are most likely out-of-distribution relative to LLaVA and Qwen's pre-training mix.
We remark a particular weakness of Qwen when prompted on \textsc{InstaOrder (Occ.)}. We noted that the model consistently answered ``no'', resulting in an extremely low F1. On the other hand, our \Ours{} are able to robustly classify these examples in the probing set. We highlight tricky cases where \Ours{} are accurate in~\cref{fig:qual_res}. Still in this figure, the A-OKVQA case shows that the model sometimes does not have a clear representation of the concept described in the image, as it replies equally for each type of pepper. On the other hand, \Ours{} manage to sharply determine which type of pepper is sold.
\input{tables/first_vs_last_token}

We investigate the quantity of probed neurons in~\cref{fig:neurons_per_layer}. Since we behave at the scalar-level and not at the vector-level anymore, we observe a large amount of accurate \Ours{}. It is especially interesting to note that these \Ours{} are also revealed in the shallowest layers of the \llm{}, suggesting that individual neurons might be involved in the decision process of the network earlier than expected~\cite{yu25multimodalllmimagetasks}.

To further attempt to understand how \Ours{} behave with respect to the predictions of the model, we plot the \ar{} curve for different \snThresh{} in the probing test of \textsc{Pope} in~\cref{fig:agreement_rate}.
We observe a clear logarithmic trend in the agreement between LLaVA and \Ours{} until they reach the same performance. Beyond this point, we observe a sharp \ar{} drop, indicating that neurons must significantly disagree with the model to obtain more accurate answers. This makes sense: the base model provides a lower bound performance if the \Ours{} only agree with it. They need to make more accurate predictions to surpass it, and therefore must disagree with it more often.

\paragraph{Main evaluation.}
\input{tables/transfer}
We then evaluate the probed \Ours{} in the validation set of our selected dataset suite in~\cref{tab:val_results}. We observe that \Ours{} largely outperform or perform competitively against the model from which they are extracted. Superior results on both \llavaSmall{} and \qwenSmall{} support the universality of our approach. Interestingly, Qwen performs extremely poorly on the occlusion understanding task, whereas our method is sufficient to raise its F1 score above that of vanilla LLaVA.
Our method continues to provide strong results for challenging \mcq{} datasets (A-OKVQA, \textsc{ScienceQa}), consistently outperforming the base model by a significant margin (\cref{tab:val_results}).

\input{figs/qual_res_full}
\paragraph{Comparison with baselines.}
We compare inference with \Ours{} against \(n\)-shot prompting in~\cref{tab:nshot_prompting}.
For both \textsc{Pope} and \textsc{Clevr}, \(n\)-shot prompting \llavaSmall{} results in large performance degradation as previously observed in~\cite{mitra25savs}. Our proposed \Ours{} outperform or obtain on par results with all the baselines for accuracy and F1.
We also compare \Ours{} with \savs{}~\cite{mitra25savs} in~\cref{tab:sns_vs_savs}. While \savs{} were originally evaluated on \textsc{VizWiz}'s ``answerable''-``unanswerable'' questions, we observed that this data was about 77\% imbalanced towards ``unanswerable'' and that \savs{} answers were biased towards it. Thus, we benchmark both \savs{} and \Ours{} on our balanced ``yes''-``no'' \textsc{VizWiz} validation set, such that the accuracy value becomes more representative of the predictive capabilities of each approach. We use the number of probing samples of \savs{} (\ie 40) for both approaches.
We observe that SAVs’ recall is effectively high but lower accuracy, indicating high answer bias. On all metrics, \Ours{} compare positively to \savs{}.

\paragraph{Runtime efficiency.}
\input{tables/n_shot}
We compare the full model and \Ours{}' inference speed in~\cref{tab:runtime_benchmark}.
By bypassing the autoregressive process, \Ours{} dramatically reduces inference runtime while maintaining performance on par with the base model. We provide an in-depth profiling using huggingface's inference pipeline in~\cref{tab:profiling} (appendix~\cref{sup:profiling}).

\paragraph{Scaling.}
\label{sec:scaling_up}
\input{tables/metric_opt}
Performance analysis confirms that \Ours{} also emerge at larger scale for both \llavaLarge{} (\cref{tab:scale_up_llava}) and \qwenLarge{} (\cref{tab:scale_up_qwen}), solidifying the scalability and universality of our approach.

\subsection{Ablation studies}
\label{sec:ablations}
\paragraph{Robustness.} We evaluate the robustness of \Ours{} with a transfer experiment in~\cref{tab:transfer}. We use the official \textsc{Pope} repository to build a \textsc{Pope-Voc} validation dataset. We probe on \textsc{Pope} built from \textsc{Coco} and evaluate on \textsc{Pope-Voc}. Although the validation distribution differs from the probing one, \Ours{} remain competitive. We also discover that \Ours{} are robust to prompt changes, providing evidence that they are not overfitting on probing data nor exploiting spurious biases from the inputs (cf. appendix \ref{sup:robustness}).

\paragraph{Token selection.}
We consider which token provides the best performance for our approach. Contrary to \savs{}~\cite{mitra25savs}, \Ours{} perform better on the first token of the sequence, opening the door to \emph{extreme early exit} by completely skipping the autoregressive process of the \llm{}. We further stop at the first layer of the model by lowering the \snThresh{} while maintaining model-level performances. In that setting, inference runs 5.10\(\times\) faster than the original model. We believe that this phenomenon arises from the larger search space of our approach (cf.~\cref{tab:search_space}).

\paragraph{Data regime.}
Although our approach is training-free, \Ours{} need to be gathered using probing data. To understand how this ties to final \Ours{} performance, we sample varying probing set sizes and plot their scores over the validation set~\cref{fig:perf_wrt_num_samples}. Overall, the more data, the better. \Ours{} surpass the base model performance when using more than 100 data samples. This is the same order of magnitude as the data required by \savs{}~\cite{mitra25savs}. Performance degrades at 5K samples, thus we default to 3K.

\paragraph{Metric optimization.}
We optimize \Ours{} for different metrics \(\mathMetric\) in table~\cref{tab:probing_results}. We then use the probed \Ours{} to validate performance in~\cref{tab:metric_opt}. This strongly influences the results on the validation data. This can be critical for sensitive applications where negative predictions can have more impact than positive (or vice-versa), such as cancer classification, where false negatives can endanger patients and erode human-machine trust.

%% file: figs/sn_per_layers.tex
\ifdefined\maketitlesupplementary
    \StartDynFig{t}{R}{0.5\linewidth}
        \centering
        \includegraphics[width=\linewidth]{assets/vizwiz/accuracy0656_t=0_layer_distribution.pdf}
        \caption{\textbf{Location of probed \Ours{} on \textsc{VizWiz}.} We use \llavaSmall{}. The \(x\)-axis indicates the layer index, while the \(y\)-axis reports how many \Ours{} were found in that layer. We use \snThresh{} \(=0.656\) to show the location of all activations that obtain a better score than the model itself.
        }
        \label{fig:neurons_per_layer}
    \EndDynFig{}
\fi

%% file: tables/probing_results.tex
\begin{table*}[t]
    \centering
    \caption{\textbf{Best probed \Our{} on diverse categorical \vqa{} datasets.} We report accuracy and F1. We display the results obtained by the single best performing \Our{} in the model. Best results are in \textbf{bold}. Baselines are highlighted in \colorbox{lightGray}{gray}. We use a probing set size of 3,000 samples except for \textsc{VizWiz}, which only contains 942 binary questions. Accuracy and F1 are optimized separately. \Ours{} outperform the base models in all cases.}
    \label{tab:probing_results}
    \resizebox{\linewidth}{!}{
    \begin{tabular}{l *{8}G}
        \toprule
        \rowcolor{white} & & \multicolumn{7}{c}{Dataset} \\
        \cmidrule[0.5pt](rl){3-9}
        \rowcolor{white}\multirow{-2}{*}{Metric} & \multirow{-2}{*}{Method} & \textsc{Pope} & \makecell{\textsc{InstaOrder} \\ \textsc{(Depth)}} & \makecell{\textsc{InstaOrder} \\ \textsc{(Occ.)}} & \textsc{VizWiz} & \textsc{Clevr} & A-OKVQA & \textsc{ScienceQA} \\
        \midrule
        \midrule
        & \llavaSmallIt{} & 90.7 & 61.7 & 53.9 & 64.8 & 52.3 & 67.8 & 62.2 \\
        \rowcolor{white} & \Our{} & \textbf{92.5\benchIncreaseFmt{1.8}} & \textbf{63.5\benchIncreaseFmt{1.8}} & \textbf{62.7\benchIncreaseFmt{8.8}} &  \textbf{71.9\benchIncreaseFmt{7.1}} & \textbf{54.4\benchIncreaseFmt{2.1}} & \textbf{68.0\benchIncreaseFmt{0.2}} & \textbf{63.8\benchIncreaseFmt{1.6}} \\
        & \qwenSmallIt{} & 95.0 & 61.8 & 50.8 & 78.3 & 84.7 & 82.7 & 81.3 \\
        \rowcolor{white}\multirow{-4}{*}{Accuracy} & \Our{} & \textbf{96.1\benchIncreaseFmt{1.1}} & \textbf{62.8\benchIncreaseFmt{1.0}} & \textbf{62.6\benchIncreaseFmt{11.8}} & \textbf{81.0\benchIncreaseFmt{2.7}} & \textbf{88.6\benchIncreaseFmt{3.9}} & \textbf{83.0\benchIncreaseFmt{0.3}} & \textbf{82.3\benchIncreaseFmt{1.0}} \\
        \midrule
        & \llavaSmallIt{} & 91.1 & 67.7 & 46.1 & 72.0 & 61.7 & \textbf{74.5} & 70.0 \\
        \rowcolor{white} & \Our{} & \textbf{92.3\benchIncreaseFmt{1.2}} & \textbf{69.1\benchIncreaseFmt{1.4}} & \textbf{69.0\benchIncreaseFmt{22.9}} & \textbf{74.8\benchIncreaseFmt{2.8}} & \textbf{66.9\benchIncreaseFmt{5.2}} & 73.8\benchDecreaseFmt{0.7} & \textbf{70.7\benchIncreaseFmt{0.7}} \\
        & \qwenSmallIt{} & 94.8 & 63.1 & 4.2 & 78.9 & 86.4 & 82.3 & 81.4 \\
        \rowcolor{white}\multirow{-4}{*}{F1} & \Our{} & \textbf{95.9\benchIncreaseFmt{1.1}} & \textbf{69.1\benchIncreaseFmt{6.0}} & \textbf{69.1\benchIncreaseFmt{64.9}} & \textbf{81.2\benchIncreaseFmt{2.3}} & \textbf{89.0\benchIncreaseFmt{2.6}} & \textbf{82.6\benchIncreaseFmt{0.3}} & \textbf{82.2\benchIncreaseFmt{0.8}} \\
        \bottomrule
    \end{tabular}
    }
\end{table*}

%% file: tables/val_results.tex
\begin{table*}[t]
    \centering
    \caption{\textbf{Evaluation of probed \Ours{} on diverse categorical VQA validation datasets.} We report both the mean of \Ours{} and the majority voting with respect to the \snThresh{} found from the probing data. All \Ours{} were optimized for accuracy on the probing set. For a given metric, best results are in \textbf{bold} and second best are \underline{underlined}. Baselines are highlighted in \colorbox{lightGray}{gray}.}
    \label{tab:val_results}
    \resizebox{\linewidth}{!}{
    \begin{tabular}{lG *{7}G}
        \toprule
        \rowcolor{white}& & \multicolumn{7}{c}{Dataset} \\
        \cmidrule[0.5pt](rl){3-9}
        \rowcolor{white}\multirow{-2}{*}{Metric} & \multirow{-2}{*}{Method} & \textsc{Pope} & \makecell{\textsc{InstaOrder} \\ \textsc{(Depth)}} & \makecell{\textsc{InstaOrder} \\ \textsc{(Occ.)}} & \textsc{VizWiz} & \textsc{Clevr} & A-OKVQA & \textsc{ScienceQA} \\
        \midrule
        \midrule
        \rowcolor{white} & \Ours{} \llavaSmallIt{} & 92\% & 63\% & 62\% & 71\% & 53\% & 67\% & 63\% \\
        \rowcolor{white}\multirow{-2}{*}{\snThresh{}} & \Ours{} \qwenSmallIt{} & 95.8\% & 62.6\% & 62.0\% & 80.9\% & 88.3\% & 80.0\% & 82.0\% \\
        \midrule
        \midrule
        & \llavaSmallIt{} & 89.8 & 65.0 & \underline{64.9} & 65.6 & 51.3 & 54.8 & \underline{53.5} \\
        \rowcolor{white} & \Ours{} (mean) & \textbf{90.9\benchIncreaseFmt{1.1}} & \textbf{66.1\benchIncreaseFmt{1.1}} & 64.5\benchDecreaseFmt{0.4} & \textbf{72.6\benchIncreaseFmt{7.0}} & \textbf{51.5\benchIncreaseFmt{0.2}} & \underline{61.4}\benchIncreaseFmt{6.6} & \textbf{57.4\benchIncreaseFmt{6.9}} \\
        \rowcolor{white} & \Ours{} (maj. voting) & \textbf{90.9\benchIncreaseFmt{1.1}} & \underline{65.2}\benchIncreaseFmt{0.2} & \textbf{78.2\benchIncreaseFmt{13.3}} & \underline{69.9}\benchIncreaseFmt{4.3} & \underline{51.4}\benchIncreaseFmt{0.1} & \textbf{72.0\benchIncreaseFmt{17.2}} & 48.4\benchDecreaseFmt{5.1} \\
        & \qwenSmallIt{} & 91.5 & 63.4 & \textbf{85.2} & \textbf{81.7} & 83.7 & \textbf{85.2} & 81.1 \\
        \rowcolor{white} & \Ours{} (mean) & \underline{93.4}\benchIncreaseFmt{1.9} & \underline{63.6}\benchIncreaseFmt{0.2} & \underline{78.1}\benchDecreaseFmt{7.1} & \underline{81.2}\benchDecreaseFmt{0.5} & \textbf{88.3\benchIncreaseFmt{4.6}} & \textbf{85.2\benchIncreaseFmt{0.0}} & \textbf{82.9\benchIncreaseFmt{1.8}} \\
        \rowcolor{white}\multirow{-6}{*}{Accuracy} & \Ours{} (maj. voting) & \textbf{93.7\benchIncreaseFmt{2.2}} & \textbf{64.9\benchIncreaseFmt{1.5}} & 77.2\benchDecreaseFmt{8.0} & 80.1\benchDecreaseFmt{1.6} & \textbf{88.3\benchIncreaseFmt{4.6}} & \textbf{85.2\benchIncreaseFmt{0.0}} & \underline{82.4}\benchIncreaseFmt{1.3} \\
        
        \midrule
        & \llavaSmallIt{} & 89.5 & \underline{66.4} & 18.5 & 60.1 & \underline{50.9} & 34.9 & \underline{40.1} \\
        \rowcolor{white} & \Ours{} (mean) & \textbf{92.6\benchIncreaseFmt{3.1}} & \textbf{68.8\benchIncreaseFmt{2.4}} & \underline{23.5}\benchIncreaseFmt{5.0} & \textbf{73.9\benchIncreaseFmt{13.9}} & \textbf{51.0\benchIncreaseFmt{1.0}} & \underline{37.7}\benchIncreaseFmt{2.8} & \textbf{42.4\benchIncreaseFmt{2.3}} \\
        \rowcolor{white} & \Ours{} (maj. voting) & \underline{91.0}\benchIncreaseFmt{0.5} & \underline{66.4}\benchIncreaseFmt{0.0} & \textbf{30.4\benchIncreaseFmt{11.9}} & \underline{64.1}\benchIncreaseFmt{4.0} & \underline{50.9}\benchIncreaseFmt{0.0} & \textbf{44.7\benchIncreaseFmt{9.8}} & 38.3\benchDecreaseFmt{1.7} \\
        & \qwenSmallIt{} & \textbf{98.5} & \underline{69.3} & \textbf{49.6} & \textbf{80.4} & 76.7 & \textbf{66.0} & 68.3 \\
        \rowcolor{white} & \Ours{} (mean) & \underline{97.6}\benchDecreaseFmt{0.9} & \textbf{70.1\benchIncreaseFmt{0.8}} & \underline{31.3}\benchDecreaseFmt{18.3} & 76.9\benchDecreaseFmt{3.5} & \textbf{87.9\benchIncreaseFmt{11.2}} & \textbf{66.0\benchIncreaseFmt{0.0}} & \underline{71.7}\benchIncreaseFmt{3.4} \\
        \rowcolor{white}\multirow{-6}{*}{Precision} & \Ours{} (maj. voting) & 96.5\benchDecreaseFmt{2.0} & 69.2\benchDecreaseFmt{0.1} & 29.2\benchDecreaseFmt{20.4} & \underline{79.8}\benchDecreaseFmt{0.6} & \underline{85.3}\benchIncreaseFmt{8.6} & \textbf{66.0}\benchIncreaseFmt{0.0} & \textbf{72.9\benchIncreaseFmt{4.6}} \\
        
        \midrule
        & \llavaSmallIt{} & \underline{90.8} & \underline{80.2} & \underline{40.2} & \textbf{92.5} & \underline{76.6} & \textbf{93.9} & \underline{85.9} \\
        \rowcolor{white} & \Ours{} (mean) & 89.5\benchDecreaseFmt{1.3} & 75.9\benchDecreaseFmt{5.3} & \textbf{61.8\benchIncreaseFmt{21.6}} & 69.9\benchDecreaseFmt{22.6} & 75.8\benchDecreaseFmt{0.8} & \underline{83.7}\benchDecreaseFmt{10.2} & 84.5\benchDecreaseFmt{1.4} \\
        \rowcolor{white}  & \Ours{} (maj. voting) & \textbf{91.4\benchIncreaseFmt{0.6}} & \textbf{80.9\benchIncreaseFmt{0.7}} & 36.5\benchDecreaseFmt{3.7} & \underline{90.3}\benchDecreaseFmt{2.2} & \textbf{77.5\benchIncreaseFmt{0.9}} & 51.8\benchDecreaseFmt{42.1} & \textbf{94.9\benchIncreaseFmt{9.0}} \\
        & \qwenSmallIt{} & 84.9 & \underline{66.0} & 2.7 & \underline{83.9} & \textbf{96.8} & \textbf{83.8} & \textbf{78.7} \\
        \rowcolor{white} & \Ours{} (mean) & \underline{89.3}\benchIncreaseFmt{4.4} & 64.8\benchDecreaseFmt{1.2} & \textbf{40.0\benchIncreaseFmt{37.3}} & \textbf{89.2\benchIncreaseFmt{5.3}} & 90.0\benchDecreaseFmt{6.8} & 83.7\benchDecreaseFmt{0.1} & \underline{78.3}\benchDecreaseFmt{0.4} \\
        \rowcolor{white}\multirow{-6}{*}{Recall} & \Ours{} (maj. voting) & \textbf{91.1\benchIncreaseFmt{6.2}} & \textbf{71.1\benchIncreaseFmt{5.1}} & \underline{37.8}\benchIncreaseFmt{35.1} & 80.6\benchDecreaseFmt{3.3} & \underline{92.5}\benchDecreaseFmt{4.3} & \textbf{83.8\benchIncreaseFmt{0.0}} & 73.4\benchIncreaseFmt{5.3} \\
        
        \midrule
        & \llavaSmallIt{} & 90.2 & \underline{72.6} & 25.3 & \underline{72.9} & \underline{61.1} & \underline{50.9} & \underline{54.7} \\
        \rowcolor{white} & \Ours{} (mean) & \underline{91.0}\benchIncreaseFmt{0.8} & 72.2\benchDecreaseFmt{0.4} & \textbf{34.0\benchIncreaseFmt{8.7}} & 71.8\benchDecreaseFmt{1.1} & 61.0\benchDecreaseFmt{0.1} & \textbf{52.0\benchIncreaseFmt{0.1}} & \textbf{56.5\benchIncreaseFmt{1.8}} \\
        \rowcolor{white} & \Ours{} (maj. voting) & \textbf{91.2\benchIncreaseFmt{1.0}} & \textbf{72.9\benchIncreaseFmt{0.3}} & \underline{33.2}\benchIncreaseFmt{7.9} & \textbf{75.0\benchIncreaseFmt{2.1}} & \textbf{61.4\benchIncreaseFmt{0.3}} & 48.0\benchDecreaseFmt{2.9} & 54.6\benchDecreaseFmt{0.1} \\
        & \qwenSmallIt{} & 91.2 & \underline{67.6} & 5.1 & \underline{82.1} & 85.6 & \underline{73.8} & 73.1 \\
        \rowcolor{white} & \Ours{} (mean) & \underline{93.3}\benchIncreaseFmt{2.1} & 67.4\benchDecreaseFmt{0.2} & \textbf{35.1\benchIncreaseFmt{30}} & \textbf{82.6\benchIncreaseFmt{0.5}} & \textbf{89.0\benchIncreaseFmt{3.4}} & \underline{73.8}\benchIncreaseFmt{0.0} & \textbf{74.8\benchIncreaseFmt{1.7}} \\
        \rowcolor{white}\multirow{-6}{*}{F1} & \Ours{} (maj. voting) & \textbf{93.7\benchIncreaseFmt{2.5}} & \textbf{70.1\benchIncreaseFmt{2.5}} & \underline{33.0}\benchIncreaseFmt{27.9} & 80.2\benchDecreaseFmt{1.9} & \underline{88.7}\benchIncreaseFmt{3.1}  & \textbf{73.9}\benchIncreaseFmt{0.1} & \underline{73.2}\benchIncreaseFmt{0.1} \\
        
        \bottomrule
    \end{tabular}
}
\end{table*}

%% file: figs/agreement_rate.tex
\StartDynFig{t}{R}{0.5\linewidth}
    \centering
    \includegraphics[width=\linewidth]{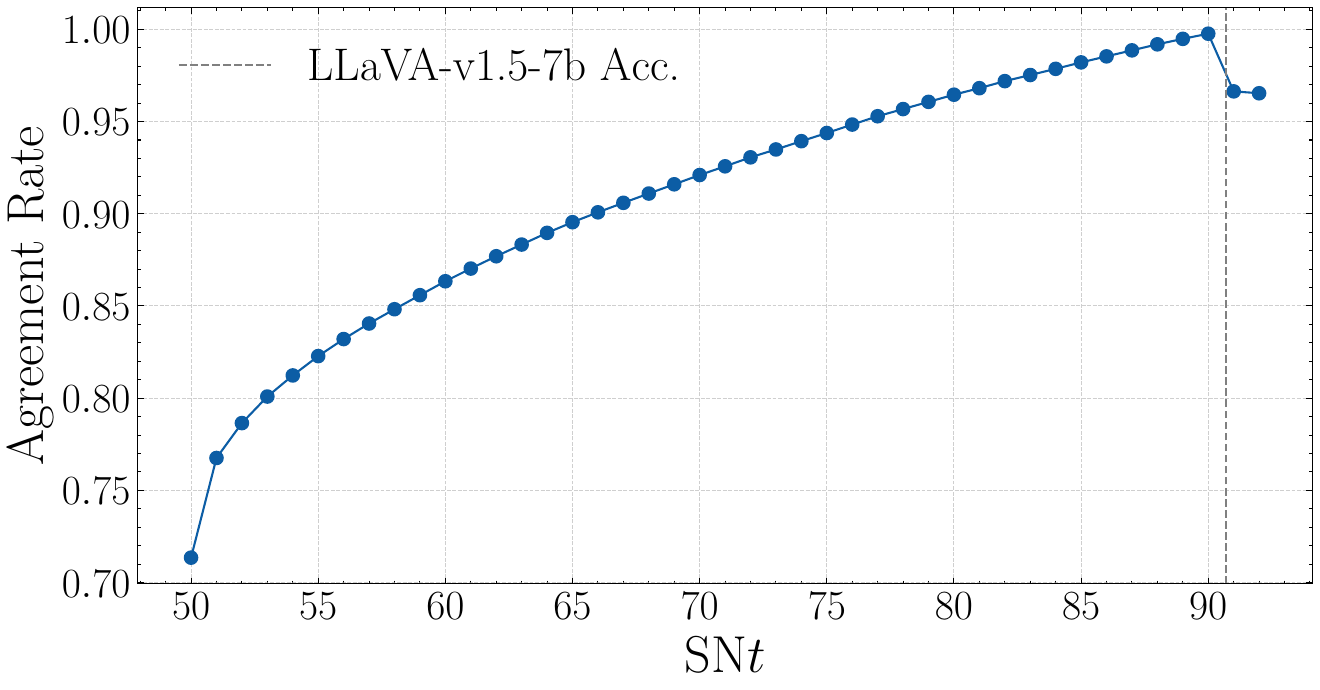}
    \caption{\textbf{\ArFull{} with respect to different \snThresh{}}. We compute \ar{} on \textsc{Pope} using \llavaSmall{}. At lower accuracy, \Ours{} largely agree with the model prediction.
    However, for \Ours{} to obtain better results than the model, they have to disagree on some answers.}
    \label{fig:agreement_rate}
\EndDynFig{}

%% file: tables/sns_vs_savs.tex
\StartDynTab{t}{R}{0.45\linewidth}
    \centering
    \caption{\textbf{Comparison of \Ours{} and \savs{}.} We benchmark \llavaSmall{} on \textsc{VizWiz} using the same number of probing samples \(N\). Best performing method is in \colorbox{lightYellow}{yellow}.}
    \label{tab:sns_vs_savs}
    \resizebox{0.95\linewidth}{!}{
    \begin{tabular}{lccccc}
         \toprule
         Method & \(\mathNumSamples\) & Acc. & Prec. & Recall & F1 \\
         \midrule
         \midrule
         SAVs~\cite{mitra25savs} & 40 & 51.8 & 50.6 & \textbf{95.7} & 66.1 \\
         \rowcolor{lightYellow}\Ours{} (ours) & 40 & \textbf{60.2} & \textbf{56.8} & 84.9 & \textbf{68.1} \\
         \bottomrule
    \end{tabular}
    }
\EndDynTab{}

%% file: tables/scale_up.tex
\begin{table*}[t]
    \caption{\textbf{Performances of \Ours{} for scaled-up models.} After probing \Ours{} following our method, we report the validation results of \llavaLarge{} and \qwenLarge{} on \textsc{Pope} and \textsc{ScienceQa} respectively. Best performances for a given optimized metric are in bold, baseline performance is in \colorbox{lightGray}{gray}.}
    \label{tab:scale_up}
    \begin{subtable}[t]{0.5\linewidth}
    \centering
    \caption{\llavaLarge{} on \textsc{Pope}.}
    \label{tab:scale_up_llava}
    \resizebox{\linewidth}{!}{
    \begin{tabular}{lGgggg}
        \toprule
        \rowcolor{white} & & \multicolumn{4}{c}{\textsc{Pope}} \\
        \cmidrule[0.5pt](rl){3-6}
        \rowcolor{white}\multirow{-2}{*}{\makecell[l]{Opt. \\ Metric}} & \multirow{-2}{*}{Method} & Acc. & Prec. & Recall & F1 \\
        \midrule
        \midrule
        & \textit{Vanilla model} & 88.7 & 85.6 & 94.0 & 89.6 \\
        \midrule
        \rowcolor{white} & \Ours{} (mean) & \textbf{88.8} & \textbf{85.7} & 94.1 & \textbf{89.7} \\
        \rowcolor{white}\multirow{-2}{*}{Accuracy} & \Ours{} (maj. voting) & 87.6 & 83.4 & \textbf{94.9} & 88.8 \\
        \midrule 
        \rowcolor{white} & \Ours{} (mean) & \textbf{88.9} & 85.5 & \textbf{94.4} & \textbf{89.7} \\
        \rowcolor{white}\multirow{-2}{*}{F1} & \Ours{} (maj. voting) & 85.9 & \textbf{85.7} & 87.2 & 86.4 \\
        \bottomrule
    \end{tabular}
    }
    \end{subtable}
    \begin{subtable}[t]{0.5\linewidth}
        \centering
        \caption{\qwenLarge{} on \textsc{ScienceQA}.}
        \label{tab:scale_up_qwen}
        \resizebox{\linewidth}{!}{
        \begin{tabular}{lGgggg}
        \toprule
        \rowcolor{white} & & \multicolumn{4}{c}{\textsc{ScienceQa}} \\
        \cmidrule[0.5pt](rl){3-6}
        \rowcolor{white}\multirow{-2}{*}{\makecell[l]{Opt. \\ Metric}} & \multirow{-2}{*}{Method} & Acc. & Prec. & Recall & F1 \\
        \midrule
        \midrule
        & \textit{Vanilla model} & 81.4 & 67.5 & 82.9 & 74.4 \\
        \midrule
        \rowcolor{white} & \Ours{} (mean) & \textbf{84.9} & 75.5 & \textbf{79.4} & \textbf{77.4} \\
        \rowcolor{white}\multirow{-2}{*}{Accuracy} & \Ours{} (maj. voting) & 84.8 & \textbf{76.6} & 77.4 & 77.0 \\
        \midrule 
        \rowcolor{white} & \Ours{} (mean) & 84.0 & 73.6 & \textbf{79.5} & 76.5 \\
        \rowcolor{white}\multirow{-2}{*}{F1} & \Ours{} (maj. voting) & \textbf{85.0} & \textbf{76.3} & 78.2 & \textbf{77.2} \\
        \bottomrule
    \end{tabular}
    }
    \end{subtable}
\end{table*}

%% file: figs/perf_wrt_num_samples.tex
\begin{figure*}[t]
    \centering
    \includegraphics[width=\linewidth]{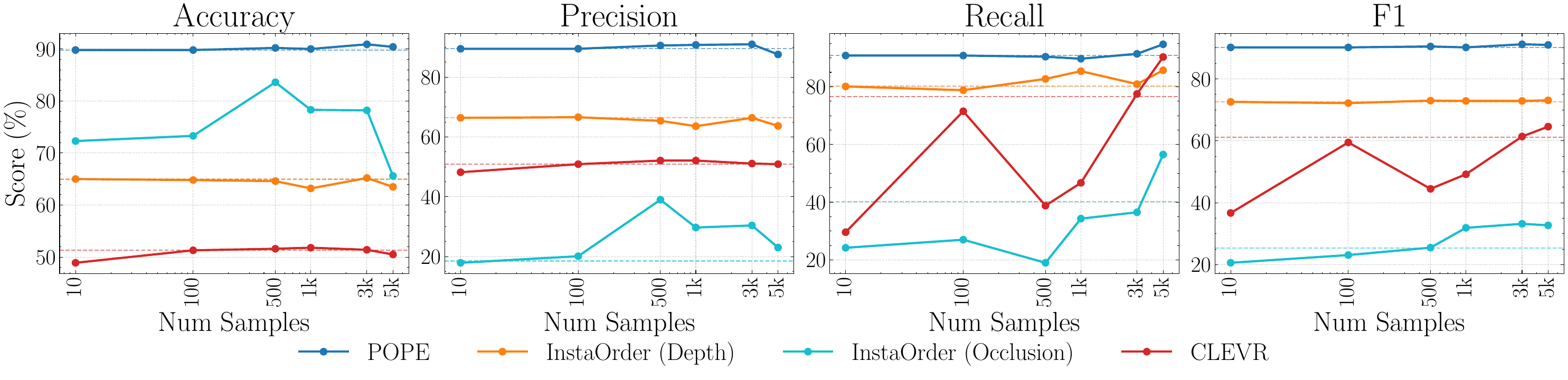}
    \caption{\textbf{Super Neurons performances with respect to different probing set sizes.} 
    We compute accuracy, precision, recall and F1 on diverse benchmarks using \llavaSmall{}. Dashed lines (colors matching their respective datasets) indicate the performance of the vanilla model. Overall, more data leads to performance improvements.}
    \label{fig:perf_wrt_num_samples}
\end{figure*}

%% file: tables/runtime.tex
\StartDynTab{t}{R}{0.6\linewidth}
    \centering
    \caption{\textbf{Runtime benchmark on \textsc{Pope}.} We benchmark the average runtime of \llavaSmall{} on the full validation set using \emph{its official inference code}. \(\mathNumLayers^\star\) indicates the inference exit layer. All numbers are recorded using a single NVIDIA RTX A6000. Best performances are in \textbf{bold}. Baseline is in \colorbox{lightGray}{gray}, and best runtime is in \colorbox{lightYellow}{yellow}.}
    \label{tab:runtime_benchmark}
    \resizebox{\linewidth}{!}{
    \begin{tabular}{l|GggggG}
         \toprule
         \rowcolor{white}\multicolumn{2}{l}{Inference Strategy} & Autoreg. & \(\mathNumLayers^\star\) & Acc. & F1 & Runtime (s.) \\
         \midrule
         \midrule
         Model & \llavaSmallIt{} & \cmark & 32 & 89.8 & 90.2 & 0.78 \\
         \midrule
         \rowcolor{white} & \(\text{\snThresh{}}=0.92\) (maj. vote) & \xmark & 15 & \textbf{90.9} & \textbf{91.2} & 0.16\benchIncrease{(-4.81\(\times\))} \\
         \rowcolor{white} & \(\text{\snThresh{}}=0.91\) (mean) & \xmark & 12 & 90.6 & 90.8 & 0.16\benchIncrease{(-4.87\(\times\))} \\
         \rowcolor{lightYellow}\cellcolor{white}\multirow{-3}{*}{\Ours{}} & \(\text{\snThresh{}}=0.90\) (maj. vote) & \xmark & \textbf{1} & 89.8 & 90.2 & \textbf{0.15\benchIncrease{(-5.10\(\times\))}} \\
         \bottomrule
    \end{tabular}
    }
\EndDynTab{}

%% file: tables/first_vs_last_token.tex
\StartDynTab{t}{R}{6cm}
    \centering
    \caption{\textbf{Impact of token selection for \Ours{} discovery on \textsc{Pope}.} We use \llavaSmall{}. \(\mathNumLayers^\star\) is the first possible early stopping layer. Best performing method is highlighted in \colorbox{lightYellow}{yellow}. Early exiting after the first token improves runtime and performances.}
    \label{tab:first_vs_last_token}
    \resizebox{\linewidth}{!}{
    \begin{tabular}{Yyyyyyl}
        \toprule
         \rowcolor{white}Token Pos. & \(\mathNumLayers^\star\) &  Acc. & Prec. & Recall & F1 & Runtime (s.) \\
         \midrule
         \midrule
         \rowcolor{white}Last & \textbf{14} & \textbf{90.9} & \textbf{92.6} & 89.5 & 91.0 & 0.76 \\
         First & 15 & \textbf{90.9} & 91.0 & \textbf{91.4} & \textbf{91.2} & \cellcolor{lightYellow}\textbf{0.16\benchIncrease{(-4.75\(\times\))}} \\
         \bottomrule
    \end{tabular}
    }
\EndDynTab{}

%% file: tables/transfer.tex
\StartDynTab{t}{R}{6cm}
    \centering
    \caption{\textbf{Transfer capabilities of \Ours{} to a novel distribution.} We use the \textsc{Pope} format.}
    \label{tab:transfer}
    \resizebox{\linewidth}{!}{
    \begin{tabular}{Ggggggg}
        \toprule
         \rowcolor{white}Method & Probe & Val. &  Acc. & Prec. & Recall & F1 \\
         \midrule
         \midrule
         LLaVA-v1.5-7b & -- & \textsc{Voc} & 91.8 & 88.1 & 96.7 & 92.2 \\
         \rowcolor{white}\Ours{} & \textsc{Voc} & \textsc{Voc} & 91.9 & 88.1 & 96.8 & 92.3 \\
         \rowcolor{lightYellow}\Ours{} & \textsc{Coco} & \textsc{Voc} & 91.6 & 87.2 & 97.6 & 92.1 \\
         \bottomrule
    \end{tabular}
    }
\EndDynTab{}

%% file: figs/qual_res_full.tex
\begin{figure*}[t]
    \centering
    \includegraphics[width=\linewidth]{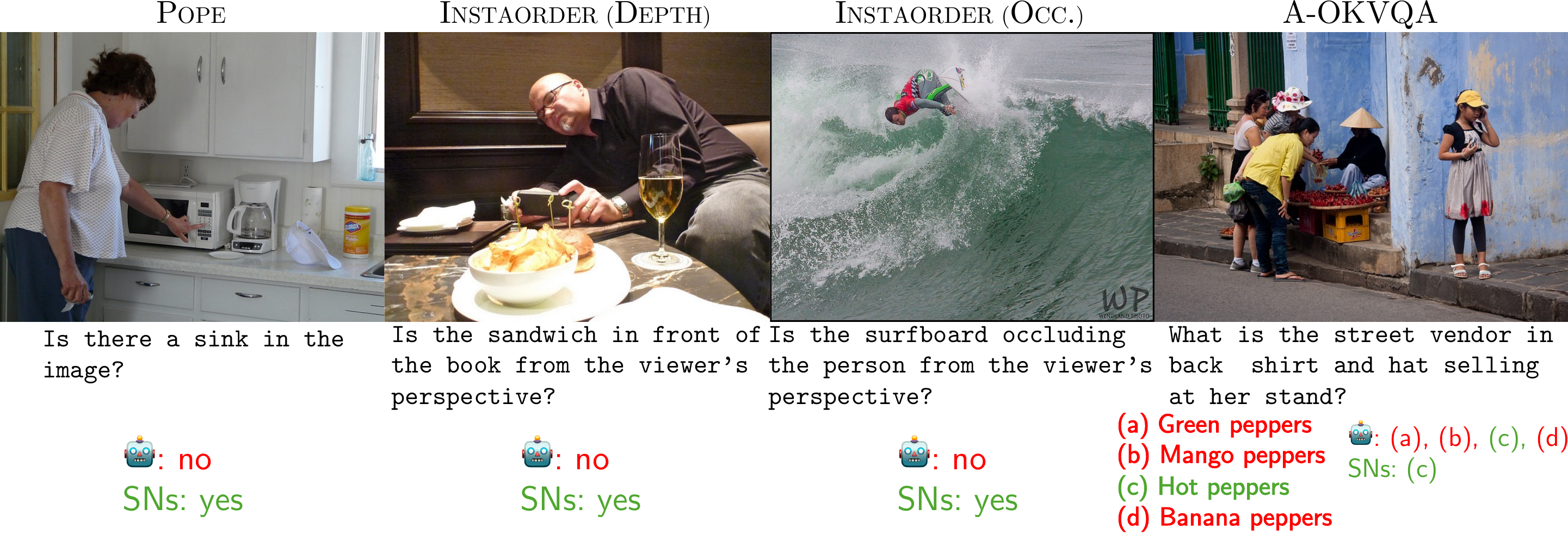}
    \caption{\textbf{Qualitative results of \Ours{}.} We show examples on diverse validation sets. We use \llavaSmall{} on \textsc{Pope} and \textsc{InstaOrder (Occ.)} and \qwenSmall{} on \textsc{InstaOrder (Depth)} and A-OKVQA. The robot icon symbolizes the model output. Green indicate that the answer matches the ground-truth and red indicates an incorrect answer.}
    \label{fig:qual_res}
\end{figure*}

%% file: tables/n_shot.tex
\StartDynTab{t}{R}{6cm}
    \caption{\textbf{Comparing \(n\)-shot prompting and \Ours{}.} We benchmark \llavaSmall{}. \Ours{} are consistently more accurate.}
    \label{tab:nshot_prompting}
    \centering
    \resizebox{0.9\linewidth}{!}{
    \begin{tabular}{c|lcccc|c}
         \toprule
         \multicolumn{2}{l}{Num. shots} & 0 & 1 & 3 & 5 & \Ours{} \\
         \midrule
         \midrule
         & Acc. & \cellcolor{lightGray}89.8 & 82.3 & 63.0 & 60.8 & \cellcolor{lightYellow}\textbf{90.9} \\ 
         & Prec. & \cellcolor{lightGray}89.5 & 84.0 & 84.0 & 81.6 & \cellcolor{lightYellow}\textbf{91.0} \\ 
         & Recall & \cellcolor{lightGray}90.8 & 81.3 & 81.3 & 31.0 & \cellcolor{lightYellow}\textbf{91.4} \\ 
         \multirow{-4}{*}{\rotatebox[origin=c]{90}{\textsc{Pope}}} & F1 & \cellcolor{lightGray}90.2 & 83.6 & 82.6 & 44.9 & \cellcolor{lightYellow}\textbf{91.2} \\ 
         \midrule
         & Acc. & \cellcolor{lightGray}65.0 & 60.7 & 58.9 & 58.9 & \cellcolor{lightYellow}\textbf{65.2} \\ 
         & Prec. & \cellcolor{lightGray}\textbf{66.4} & 60.3 & 58.6 & 58.7 & \cellcolor{lightYellow}\textbf{66.4} \\ 
         & Recall & \cellcolor{lightGray}80.2 & 93.9 & \textbf{98.7} & 97.9 & \cellcolor{lightYellow}80.9 \\ 
         \multirow{-4}{*}{\rotatebox[origin=c]{90}{\makecell{\scriptsize{\textsc{InstaOrder}} \\ \scriptsize{\textsc{(Depth)}}}}} & F1 & \cellcolor{lightGray}72.6 & \textbf{73.5} & \textbf{73.5} & 73.4 & \cellcolor{lightYellow}72.9\\ 
         \midrule
         & Acc. & \cellcolor{lightGray}51.3 & 50.8 & 50.4 & 49.9 & \cellcolor{lightYellow}\textbf{51.4} \\ 
         & Prec. & \cellcolor{lightGray}\textbf{50.9} & 50.7 & 50.4 & 49.9 & \cellcolor{lightYellow}\textbf{50.9} \\ 
         & Recall & \cellcolor{lightGray}76.6 & 59.2 & 51.2 & 50.7 & \cellcolor{lightYellow}\textbf{77.5} \\ 
         \multirow{-4}{*}{\rotatebox[origin=c]{90}{\textsc{Clevr}}} & F1 & \cellcolor{lightGray}61.1 & 54.6 & 50.8 & 50.3 & \cellcolor{lightYellow}\textbf{61.4}\\ 
         \bottomrule
    \end{tabular}
    }
\EndDynTab{}

%% file: tables/metric_opt.tex
\StartDynTab{t}{R}{6cm}
    \centering
    \caption{\textbf{Impact of metric optimization for \Ours{} discovery.} We use \llavaSmall{} on \textsc{InstOrder (Occ.)}. Willingly optimizing for a given metric \(\mathMetric\) results in significant performance improvements in the subsequent measure.}
    \label{tab:metric_opt}
    \resizebox{\linewidth}{!}{
    \begin{tabular}{llcc}
        \toprule
        \rowcolor{white} & & \multicolumn{2}{c}{Metric optimization (\(\mathMetric\))} \\
        \cmidrule[0.5pt](rl){3-4}
        \rowcolor{white}\multirow{-2}{*}{Metric} & \multirow{-2}{*}{Method} & Accuracy & F1 \\
        \midrule
        \midrule
        \rowcolor{white} & \Ours{} (mean) & 64.5 & 25.6 \\
        \rowcolor{white}\multirow{-2}{*}{Accuracy} & \Ours{} (maj. voting) & \textbf{78.2} & \textbf{67.0} \\
        \midrule
        \rowcolor{white} & \Ours{} (mean) & 23.5 & 16.4 \\
        \rowcolor{white}\multirow{-2}{*}{Precision} & \Ours{} (maj. voting) & \textbf{30.4} & \textbf{24.6} \\
        \midrule
        \rowcolor{white} & \Ours{} (mean) & \textbf{61.8} & \textbf{98.4} \\
        \rowcolor{white}\multirow{-2}{*}{Recall} & \Ours{} (maj. voting) & 36.5 & 59.7  \\
        \midrule
        \rowcolor{white} & \Ours{} (mean) & \textbf{34.0} & 28.1  \\
        \rowcolor{white}\multirow{-2}{*}{F1} & \Ours{} (maj. voting) & 33.2 & \textbf{34.9} \\
        \bottomrule
    \end{tabular}
    }
\EndDynTab{}

%% file: sec/5_conclusion.tex
\section{Conclusion}
\label{sec:conclusion}

We highlight that scalar activations deemed \OursFull{} can serve as strong and robust categorical classifiers while improving runtime efficiency of \vlms{}. \Ours{} compare favorably with the base models they are extracted from and \savsFull{}, while enabling discovery of a large amount of accurate scalars in shallower layers of the \llm{}. This enables \emph{extreme early exiting} during the generation of the \emph{first token} and in the \emph{first layer} of the \llm{}, substantially reducing inference time. We plan to apply our method for vision language action models, in which accurate discrete action decisions could be taken faster.

\paragraph{Limitations.}
Future research should be conducted to determine whether \OursFull{} are able to provide useful signals on complex open-ended prompts and reasoning and if robustness and early exiting still hold in these contexts.

%% file: sec/supp/0_suppl.tex
\clearpage
\setcounter{page}{1}

\ifdefined\maketitlesupplementary
  \maketitlesupplementary
\fi

\title{\paperTitle{} \\ \normalfont \textit{Supplementary Materials}}
\author{\paperAuthors{}}
\authorrunning{\runningAuthors{}}
\institute{\paperInstitute{}}
\titlerunning{\paperTitleRunning}

\renewcommand{\theHsection}{A\arabic{section}}

\appendix  
\maketitle

\input{sec/supp/A_datasets}
\input{sec/supp/B_baselines}
\input{sec/supp/C_profiling}
\input{sec/supp/D_robustness}
\input{sec/supp/E_sn_location}

%% file: sec/supp/A_datasets.tex
\input{tables/prompt_template}
\section{Datasets and prompts}
\label{sup:datasets_and_prompts}
We describe the dataset along with the prompts used for probing and validation in more detail in this section.

\paragraph{\textsc{Pope}.}
\textsc{Pope} is a dataset designed to evaluate the degree of hallucinations of \vlms{}. Built on top of \textsc{Coco}, it consists of 2.9K \vqa{} that relate to the presence or absence of objects in images. Broadly speaking, to visual hallucinations~\cite{li2023pope}.

\paragraph{\textsc{InstaOrder (Occ.)}.}
\textsc{InstaOrder} also stems from \textsc{Coco} and consists of instance-wise ordering annotations. Recent work converted this dataset into a \bvqa{} format and observed that \llavaSmall{} struggles to answer such geometry-related prompts. This split of the dataset tests the models' understanding of instance-wise occlusions~\cite{lee22instaorder, musacchio2025instaformer}.

\paragraph{\textsc{InstaOrder (Depth)}.}
Second split of \textsc{InstaOrder}, the questions contained in this challenge test the models' abilities to reason about instance-wise depth order.

\paragraph{\textsc{VizWiz}.}
Created by people with visual impairments, \textsc{VizWiz} is a general benchmark aiming at testing models' capacities to reason under constrained visual inputs. We filter the dataset on ``yes''-``no'' answers and only keep such questions for both the probing and validation sets~\cite{gurari2018vizwiz}.

\paragraph{\textsc{Clevr}.}
To test our method's generalization capabilities on synthetic data, we select \textsc{Clevr}. This dataset evaluates the spatial understanding of the model through questions that span from counting to positional understanding. We filter this dataset to only retain classification questions for probing and validation~\cite{johnson2017clevr}.

\paragraph{A-OKVQA.} A-OKVQA is an \mcq{} dataset of with four answer choices consisting of general visual world knowledge. Questions require commonsense reasoning grounded in the scene in order to be accurately answered~\cite{schwenk2022aokvqa}.

\paragraph{\textsc{ScienceQa.}} We benchmark our approach on this high school science dataset. The \textsc{ScienceQa} dataset contains 26 topics, 127 categories, and 379 skills~\cite{lu22scienceqa}. We convert this \mcq{} data to a series of \bvqaFull{} and perform inference in the same way as on the other datasets.

\paragraph{Prompts.} For \textsc{Clevr} and \textsc{VizWiz}, we do not apply any prompt template and simply filter the dataset to retain only ``yes''-``no'' \vqa{} in both the training and validation sets. We prompt the model with the filtered samples. For the other datasets, we report the prompt templates we use in~\cref{tab:prompt_template}.

\input{tables/config_models}
\input{tables/config_llava}
\input{tables/config_qwen}

Note that for \textsc{InstaOrder}, the location of the object is only specified if there are multiple same object categories in the image. This allows disambiguation of the target object. Otherwise, only the object category is specified.

%% file: tables/prompt_template.tex
\begin{table*}[t]
    \centering
    \caption{\textbf{Prompt templates.} We detail the templates used for some of the \vqa{} datasets. \{\} indicates variables, \texttt{obj} indicates the object category, \texttt{loc} indicates the location of the object in a bounding box format and. For \mcq{} data, \texttt{q} indicates the original question and \(a_i\) indicates the \(i\)-th \mcq{} answer.}
    \label{tab:prompt_template}
    \begin{tabularx}{\textwidth}{llX}
        \toprule
         Dataset & Ref.~~~ & Prompt  \\
         \midrule
         \midrule
         \textsc{Pope} & \cite{li2023pope} & \texttt{Is there a \{obj\} in the image?} \\
         \textsc{InstaOrder (Occ.)} & \cite{musacchio2025instaformer} & \texttt{Is the \{obj\} at position \{loc\} obstructing the \{obj\} at position \{loc\}? Answer the question using a single word.} \\
         \textsc{InstaOrder (Depth.)} & \cite{musacchio2025instaformer} & \texttt{Is the \{obj\} in front of the \{obj\} at position \{loc\} from the viewer's perspective? Answer the question using a single word.} \\
         A-OKVQA & \cite{schwenk2022aokvqa} & \texttt{\{q\} Is it \{\(a_i\)\}? Answer with yes or no.} \\
         \textsc{ScienceQa} & \cite{lu22scienceqa} & \texttt{\{q\} Is it \{\(a_i\)\}? Answer with yes or no.} \\
        \bottomrule
    \end{tabularx}
\end{table*}

%% file: tables/config_models.tex
\ifdefined\maketitlesupplementary
\else
\begin{table}[t]
        \caption{Default configurations of the \llavaSmall{} and \qwenSmall{} we use for all the experiments throughout our work.}
        \begin{subtable}[t]{0.5\linewidth}
            \centering
            \caption{\llavaSmall{} configuration.}
            \label{tab:config_llava}
            \begin{tabular}{ll}
            \toprule
            dtype & float16 \\
            model\_type & llava \\
            head\_dim & 128 \\
            hidden\_act & silu \\
            hidden\_size & 4096 \\
            model\_type & llama \\
            num\_attention\_heads & 32 \\
            num\_hidden\_layers & 32 \\
            num\_key\_value\_heads & 32 \\
            vocab\_size & 32064 \\
            \bottomrule
            \end{tabular}
        \end{subtable}
        \begin{subtable}[t]{0.5\linewidth}
            \centering
            \caption{\qwenSmall{} configuration.}
            \label{tab:config_qwen}
            \begin{tabular}{ll}
            \toprule
            dtype & float16 \\
            model\_type & qwen3\_vl \\
            head\_dim & 128 \\
            hidden\_act & silu \\
            hidden\_size & 2560 \\
            model\_type & qwen3\_vl\_text \\
            num\_attention\_heads & 32 \\
            num\_hidden\_layers & 36 \\
            num\_key\_value\_heads & 8 \\
            vocab\_size & 151936 \\
            \bottomrule
            \end{tabular}
        \end{subtable}
\end{table}
\fi

%% file: tables/config_llava.tex
\ifdefined\maketitlesupplementary
\begin{table}[t]
    \centering
    \caption{Configuration of \llavaSmall{} used in our experiments.}
    \label{tab:config_llava}
    \begin{tabular}{ll}
    \toprule
    dtype & float16 \\
    model\_type & llava \\
    head\_dim & 128 \\
    hidden\_act & silu \\
    hidden\_size & 4096 \\
    model\_type & llama \\
    num\_attention\_heads & 32 \\
    num\_hidden\_layers & 32 \\
    num\_key\_value\_heads & 32 \\
    vocab\_size & 32064 \\
    \bottomrule
    \end{tabular}
\end{table}
\fi

%% file: tables/config_qwen.tex
\ifdefined\maketitlesupplementary
\begin{table}[t]
    \centering
    \caption{Configuration of \qwenSmall{} used in our experiments.}
    \label{tab:config_qwen}
    \begin{tabular}{ll}
    \toprule
    dtype & float16 \\
    model\_type & qwen3\_vl \\
    head\_dim & 128 \\
    hidden\_act & silu \\
    hidden\_size & 2560 \\
    model\_type & qwen3\_vl\_text \\
    num\_attention\_heads & 32 \\
    num\_hidden\_layers & 36 \\
    num\_key\_value\_heads & 8 \\
    vocab\_size & 151936 \\
    \bottomrule
    \end{tabular}
\end{table}
\fi

%% file: sec/supp/B_baselines.tex
\input{tables/sns_vs_finetuning}

\section{Baselines}
\label{sup:baseline_comparison}

\paragraph{Baseline configurations.} We report the configurations of \llavaSmall{} and \qwenSmall{} used in our experiments in~\cref{tab:config_llava} and~\cref{tab:config_qwen}. We do not alter these configurations unless specified otherwise in the paper.

\paragraph{Finetuned baselines comparison.}
We compare \OursFull{} to fully finetuned and LoRA-tuned~\cite{hu2022lora} LLaVA~\cite{liu23llava} at different dataset scales in~\cref{tab:sup_sns_vs_finetuning}. We use a dataset on which LLaVA is known to struggle to clearly identify the data scale at which finetuning begins to outperform \Ours{}. For a small annotated finetuning budget, \Ours{} obtain better results than finetuned alternatives while being training-free. On the other hand, as the annotation budget grows, fully finetuning the model becomes a better alternative if the goal is to improve performance for a \emph{single task}. In fact, finetuning the model on a single task can lead to significant degradation on other tasks. In contrast, our training-free approach maintains the model's general capabilities by preserving the pretrained weights.

%% file: tables/sns_vs_finetuning.tex
\begin{table*}[t]
    \centering
    \caption{\textbf{\Ours{} compared to finetuned LLaVA alternatives on \textsc{InstaOrder}.} Given an annotation budget \(\mathNumSamples\), we compare both training time and task performances. We use the pretrained model of~\cite{liu23llava} as our baseline and their official finetuning script without altering any of the hyperparameters. For our approach, we report the results from the method obtaining the best F1 based on~\cref{tab:val_results}. Best results are in \textbf{bold}. Baseline is highlighted in \colorbox{lightGray}{gray}, while our \oursFull{} are in \colorbox{lightYellow}{yellow}.}
    \label{tab:sup_sns_vs_finetuning}
    \resizebox{\linewidth}{!}{
    \begin{tabular}{GGGggggggggg}
         \toprule
         \rowcolor{white}& & & & \multicolumn{4}{c}{Occlusion performances} & \multicolumn{4}{c}{Depth performances} \\
         \cmidrule[0.5pt](rl){5-8}
         \cmidrule[0.5pt](rl){9-12}
         \rowcolor{white}Method & \(\mathNumSamples\) & Training time & \makecell{Data\\generalization} & Acc. & Precision & Recall & F1 & Acc. & Precision & Recall & F1 \\
         \midrule
         \midrule
         \llavaSmallIt{} & 753k & 8 h.~\cite{liu23llava} & \xmark & 64.9 & 18.5 & 40.2 & 25.3 & 65.0 & 66.4 & 80.2 & 72.6 \\
         \midrule
         \rowcolor{white}\llavaSmallIt{}-sft & \textbf{3k} & 2m. & \xmark & 14.8 & 14.8 & \textbf{100.0} & 25.8 & 57.9 & 57.9 & \textbf{100.0} & 73.4 \\
         \rowcolor{white}\llavaSmallIt{}-sft & 10k & 5m. & \xmark & 82.1 & 43.5 & 71.2 & 54.0 & 75.1 & 83.1 & 71.5 & 76.9  \\
         \rowcolor{white}\llavaSmallIt{}-sft & 100k & 49m. & \xmark & 83.7 & 47.1 & 81.8 & 59.8 & 79.8 & 83.8 & 80.8 & 82.3 \\
         \midrule
         \rowcolor{white}\llavaSmallIt{}-LoRA & 3k & 2m. & \xmark & 67.5 & 23.9 & 54.7 & 33.3 &  55.4 & 58.8 & 76.4 & 66.5  \\
         \rowcolor{white}\llavaSmallIt{}-LoRA & 10k & 5m. & \xmark & 83.0 & 45.2 & 69.8 & 54.8 &  75.1 & 84.8 & 69.5 & 76.4 \\
         \rowcolor{white}\llavaSmallIt{}-LoRA & 100k & 41m. & \xmark & \textbf{84.4} & \textbf{48.5} & 81.2 & \textbf{60.7} & \textbf{80.2} & \textbf{84.1} & 81.3 & \textbf{82.7} \\
         \midrule
         \rowcolor{lightYellow}\OursFull{} & \textbf{3k} & \textbf{0m.} & \cmark & 64.5 & 23.5 & 61.8 & 34.0 & 65.2 & 66.4 & 80.9 & 72.9 \\
         \bottomrule
    \end{tabular}
    }
\end{table*}

%% file: sec/supp/C_profiling.tex
\section{Profiling}
\label{sup:profiling}
\input{tables/profiling}
We run an in-depth profiling benchmark using finer-grained measurements on an NVIDIA A100 GPU while varying the maximum generated token of the model (\cref{tab:profiling}). We use \llavaSmall{} from the huggingface library. \Ours{}, by skipping the autoregressive process of the network, dramatically decreases the wall time to obtain an answer. Most of the gain comes from bypassing (i) the autoregressive nature of the transformer, and (ii) avoiding the huggingface's post-processing routine. Our approach is still 1.8\(\times\) faster than the base model when capping the model’s maximum number of generated tokens at 1.

%% file: tables/profiling.tex
\StartDynTab{t}{R}{0.6\linewidth}
    \centering
    \caption{Profiling \llavaSmall{} on an NVIDIA A100 GPU using the huggingface's inference routine (in s.). Ours is benchmarked on the first layer of the \llm{}.
    }
    \label{tab:profiling}
    \resizebox{\linewidth}{!}{
    \begin{tabular}{llcccl}
        \toprule
         \rowcolor{white}Model & Max new toks. & Embedding & Prefill & Decoding & Wall time \\
         \midrule
         \midrule
         LLaVA-v1.5-7b & 128 & 0.032 & 0.085 & 0.025 & 1.01 \\
         LLaVA-v1.5-7b & 1 & 0.032 & 0.086 & 0.024 & 0.223 \\
         \rowcolor{lightYellow}\Ours{} & 1 & 0.032 & 0.085 & \textbf{0.002} & \textbf{0.119}\benchIncrease{(-1.9\(\times\))} \\
         \bottomrule
    \end{tabular}
    }
\EndDynTab{}

%% file: sec/supp/D_robustness.tex
\input{tables/prompt_robustness}
\section{Robustness}
\label{sup:robustness}
\paragraph{Prompt sensitivity.}
After investigating the robustness to dataset transfer in~\cref{sec:ablations}, we investigate the sensitivity of \Ours{} to prompts. On the \textsc{InstaOrder (Occ.)} dataset, the prompt template involves a relational word between the two compared instances (cf.~\cref{tab:prompt_template}). We replace that relational word with alternatives and report the results in~\cref{tab:prompt_robustness}. We find that \Ours{} are robust to the choice of initial prompt, as shown by their similar F1 scores. 

To push this idea, we also create a dataset that consists of random comparative strings formed by sampling 3-10 characters at random. Interestingly, \Ours{} perform better than the vanilla model on this set as well suggesting that \Ours{} encode general in-domain knowledge for a given task. This is due to the fact that we are directly extracting \Ours{} from a metric that evaluates the task of interest.

\paragraph{Adversarial prompting.}
On the other hand, these numbers might hint that \Ours{} could be overfitting or exploiting spurious bias from input data to produce their answers. To verify this, we lead two experiments using \llavaSmall{} on \textsc{Pope}. We choose \textsc{Pope} since performance on this dataset is already saturated and revolves around questions related to class recognition that can be subject to dataset class imbalance, allowing us to better perceive potential overfitting. We create two adversarial datasets:
\begin{itemize}
    \item \textsc{Pope}-\texttt{Im}: we shuffle all the images of the validation set of \textsc{Pope},
    \item \textsc{Pope}-\texttt{Txt}: we shuffle all the prompts of the validation set of \textsc{Pope}.
\end{itemize}
We evaluate \Ours{} on these datasets in~\cref{tab:pope_adversarial}. Results show that when the prompt is not grounded in the image (and vice versa), \Ours{} fail to answer accurately, and the F1 score collapses. This indicates that while the prompt must have a degree of relevance to the image, \Ours{} are likely not leveraging spurious biases or overfitting to either the prompt or the image to come up with their decisions. This claim is emphasized by the fact that \Ours{} can generalize to novel distributions as shown in~\cref{tab:transfer} of the main paper.

\input{tables/pope_adversarial}

%% file: tables/prompt_robustness.tex
\begin{table*}[t]
    \centering
    \caption{\textbf{Prompt robustness.} We use \llavaSmall{} on \textsc{InstaOrder (Occ.)}. We replace the comparative word in the prompt template with alternative words and evaluate compare the base model performances to the \Ours{} performances.}
    \label{tab:prompt_robustness}
    \begin{tabular}{llcccccccc}
        \toprule
         \multicolumn{2}{c}{Comparative word} & \multicolumn{2}{c}{\tt{obstructing}} & \multicolumn{2}{c}{\tt{hiding}} & \multicolumn{2}{c}{\tt{occluding}} & \multicolumn{2}{c}{Random} \\
         \cmidrule[0.5pt](rl){3-4}
         \cmidrule[0.5pt](rl){5-6}
         \cmidrule[0.5pt](rl){7-8}
         \cmidrule[0.5pt](rl){9-10}
         Split & Metric & Base & \Ours{} & Base & \Ours{} & Base & \Ours{} & Base & \Ours{} \\
         \midrule
         \midrule
         \multirow{2}{*}{Probe} & Acc. & 53.9 & 62.7	& 45.6 & 62.6 & 48.8 & 64.2 & 49.5 & 60.0 \\
         & F1 & 46.1	& 69.0 & 61.6 & 67.9 & 64.3 & 67.6 & 65.0 & 67.0 \\
         \midrule
         \multirow{4}{*}{Validation} & Acc. & 64.9 & 78.2 & 16.2 & 68.1 & 18.3 & 66.8 & 17.9 & 66.9 \\
         & Prec. & 18.5 & 30.4 & 13.9 & 24.2 & 14.6 & 23.8 & 14.7 & 22.2 \\
         & Recall & 40.2	& 36.5 & 89.8 & 54.1 & 92.8 & 56.5 & 94.3 & 49.2 \\
         & F1 & 25.3	& 33.2 & 24.1 & 33.5 & 25.2 & 33.5 & 25.4 & 30.6 \\
         \bottomrule
    \end{tabular}
\end{table*}

%% file: tables/pope_adversarial.tex
\begin{table*}[t]
    \centering
    \caption{\textbf{Adversarial prompting.} We evaluate \llavaSmall{} on adversarial \textsc{Pope}-\texttt{Im} and \textsc{Pope}-\texttt{Txt} validation datasets containing shuffled \vqa{} pairs. \Ours{} results on the original dataset is highlighted in \colorbox{lightGray}{gray}.}
    \label{tab:pope_adversarial}
    \begin{tabular}{lccc|c}
        \toprule
        \rowcolor{white}& & \multicolumn{3}{c}{Dataset} \\
        \cmidrule[0.5pt](rl){3-5}
        \rowcolor{white}\multirow{-2}{*}{Metric} & \multirow{-2}{*}{Method} & \textsc{Pope}-\texttt{Im} & \textsc{Pope}-\texttt{Txt} & \textsc{Pope} \\
        \midrule
        \midrule
        \rowcolor{white} & \Ours{} (mean) & 60.4 & 60.4 & \cellcolor{lightGray}\textbf{90.9} \\
        \rowcolor{white}\multirow{-2}{*}{Accuracy} & \Ours{} (maj. voting) & 60.3 & 61.6 & \cellcolor{lightGray}\textbf{90.9} \\
        
        \midrule
        \rowcolor{white} & \Ours{} (mean) & 71.2 & 71.4 & \cellcolor{lightGray}\textbf{92.6} \\
        \rowcolor{white}\multirow{-2}{*}{Precision} & \Ours{} (maj. voting) & 68.2 & 69.9 & \cellcolor{lightGray}\textbf{91.0} \\
        
        \midrule
        \rowcolor{white}& \Ours{} (mean) & 38.9 & 38.9 & \cellcolor{lightGray}\textbf{89.5} \\
        \rowcolor{white}\multirow{-2}{*}{Recall} & \Ours{} (maj. voting) & 43.2 & 44.9 & \cellcolor{lightGray}\textbf{91.4} \\
        
        \midrule
        \rowcolor{white} & \Ours{} (mean) & 50.3 & 50.3 & \cellcolor{lightGray}\textbf{91.0} \\
        \rowcolor{white}\multirow{-2}{*}{F1} & \Ours{} (maj. voting) & 52.9 & 54.6 & \cellcolor{lightGray}\textbf{91.2} \\
        \bottomrule
    \end{tabular}
\end{table*}

%% file: sec/supp/E_sn_location.tex
\section{Location of \Ours{}}

\input{figs/sns_location}

\paragraph{Dataset-wise.} We visualize the location of \Ours{} for different datasets in~\cref{fig:sns_location}. We use \llavaSmall{} and only report \Ours{} that perform better than the network itself. In~\cref{fig:sns_location_pope}, we observe that a few \Ours{} perform better than the model. We hypothesize that this is because the model already answers \textsc{Pope} correctly. Yet, results in~\cref{tab:val_results} show that these \Ours{} are still useful for improving the model's performance while dramatically reducing runtime~\cref{tab:runtime_benchmark}.

On the other hand, we observe that datasets for which \llavaSmall{} struggles naturally result in the emergence of many more \Ours{} (\cref{fig:sns_location_vizwiz}). Interestingly, a lot of them appear to be located in the shallower layers of the model.

\paragraph{Cross-dataset.} To observe if some neurons share multiple types of expertise, we plot a heatmap of overlapping \Ours{} on different datasets in~\cref{fig:overlap_sns}. We only consider the neurons that exceed the model's performance on \emph{all} of the indicated datasets. We observe that some neurons indeed capture useful information across multiple tasks, suggesting they are part of an underlying decision process within the network. Moreover, this allows us to hypothesize potential links between different types of questions. Such a heatmap offers a new lens for understanding task relations and task proximity. In the future, we plan to explore the correlation between task proximity and \Ours{} overlap.

Of particular interest, we note that \textsc{InstaOrder (Occ.)} and \textsc{Clevr} share the largest number of \Ours{}. This can be explained by the similarity between the questions in the two datasets that both target object-wise geometric understanding (\cref{fig:ovl_instaorder_occ_clevr}), but also because the base \llavaSmall{} does not perform well on these benchmarks. We find cases of \Ours{} that surpass the model's performance on three datasets at the same time (\cref{fig:ovl_instaorder_occ_instaorder_depth_clevr}) and a surprising case where a single neuron obtains better performances than \llavaSmall{} on all four benchmarks in~\cref{fig:ovl_instaorder_occ_instaorder_depth_vizwiz_clevr}. This indicates that sparse activations in the form of \OursFull{} constitute a lightweight and powerful representation that can generalize across diverse types of data.
\input{figs/overlapping_sns}

%% file: figs/sns_location.tex
\begin{figure*}
     \begin{subfigure}[b]{0.49\textwidth}
         \centering
         \includegraphics[width=\textwidth]{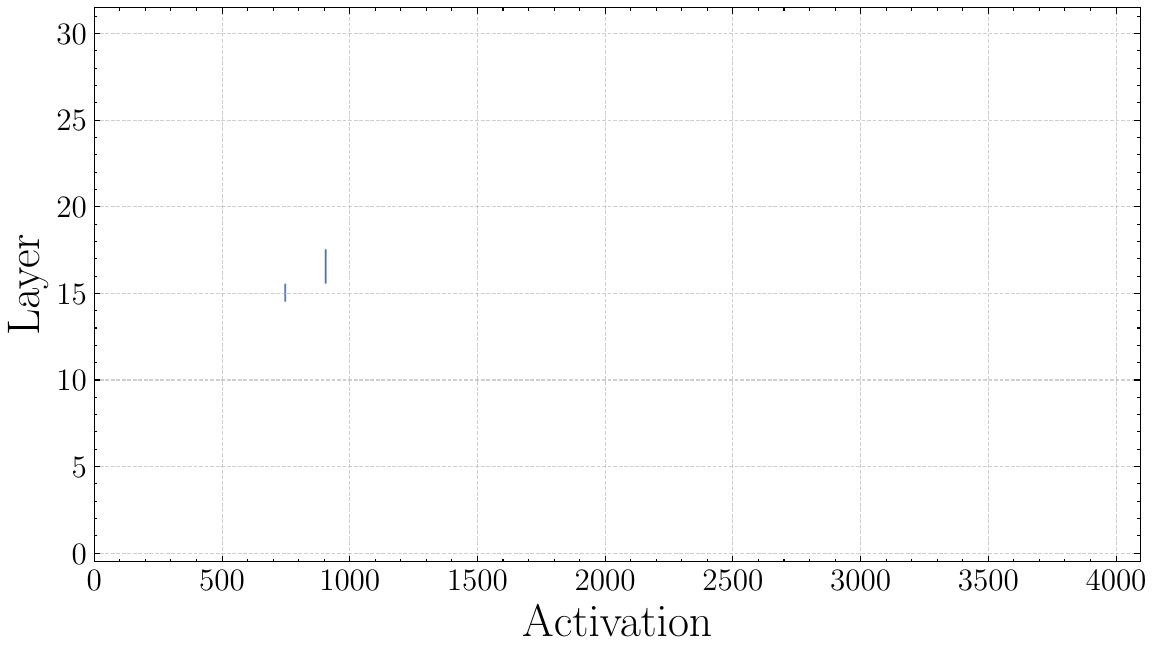}
         \caption{Location of \Ours{} on \textsc{Pope}.}
         \label{fig:sns_location_pope}
     \end{subfigure}
     \hfill
     \begin{subfigure}[b]{0.49\textwidth}
         \centering
         \includegraphics[width=\textwidth]{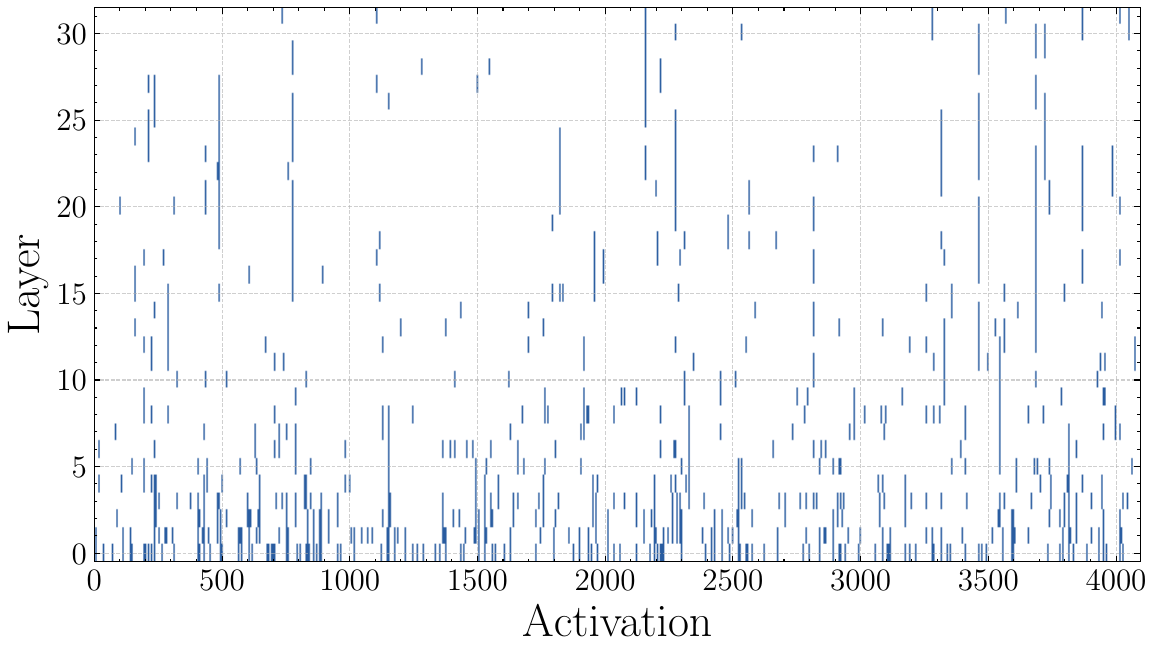}
         \caption{Location of \Ours{} on \textsc{Vizwiz}.}
         \label{fig:sns_location_vizwiz}
     \end{subfigure}
        \caption{\textbf{Visualization of \Ours{} on different datasets.} We use \llavaSmall{}. We only visualize \Ours{} that obtain better performances than the model itself. Few \Ours{} emerge on the datasets that are already very well answered by the model, while we observe a significant amount of \Ours{} on harder datasets.}
        \label{fig:sns_location}
\end{figure*}

%% file: figs/overlapping_sns.tex
\begin{figure*}
     \centering
     \begin{subfigure}[t]{0.49\textwidth}
         \centering
         \includegraphics[width=\textwidth]{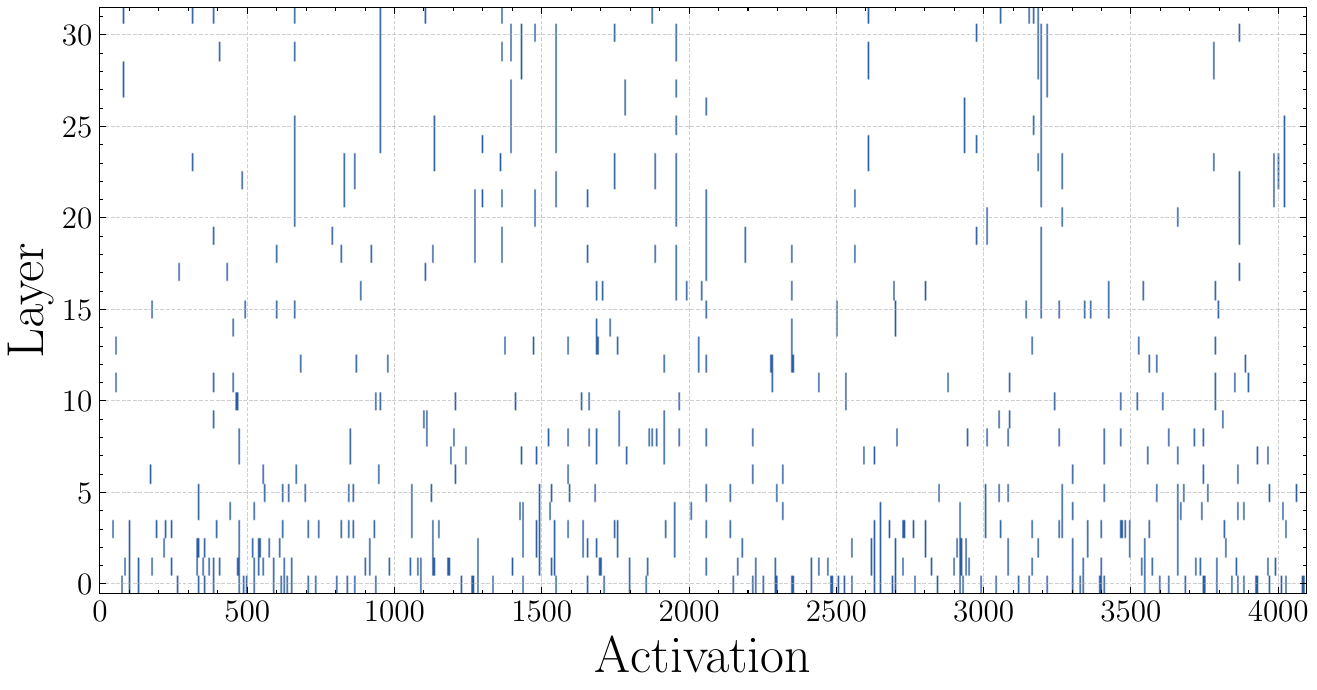}
         \caption{Overlapping \Ours{} between \textsc{Vizwiz} and \textsc{Clevr}.}
         \label{fig:ovl_vizwiz_clevr}
     \end{subfigure}
     \hfill
     \hfill
     \begin{subfigure}[t]{0.49\textwidth}
         \centering
         \includegraphics[width=\textwidth]{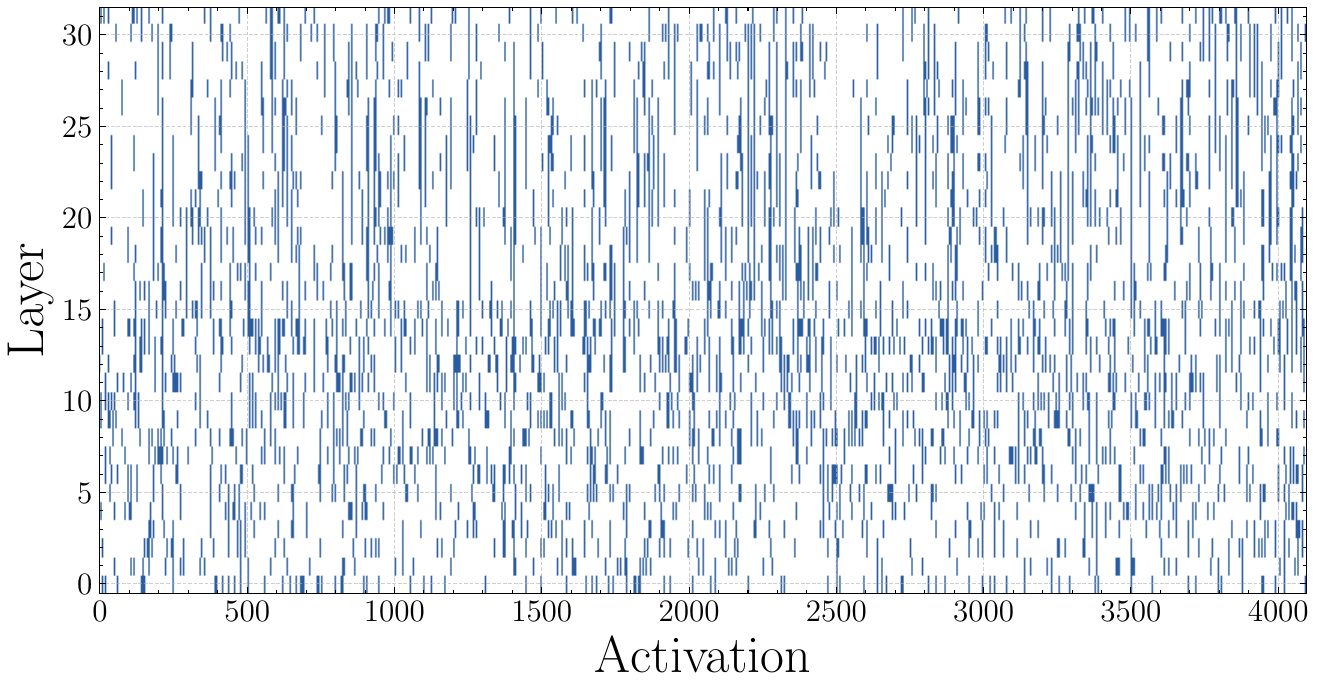}
         \caption{Overlapping \Ours{} between \textsc{InstaOrder (Occ.)} and \textsc{Clevr}.}
         \label{fig:ovl_instaorder_occ_clevr}
     \end{subfigure}
     \hfill
     \begin{subfigure}[t]{0.49\textwidth}
         \centering
         \includegraphics[width=\textwidth]{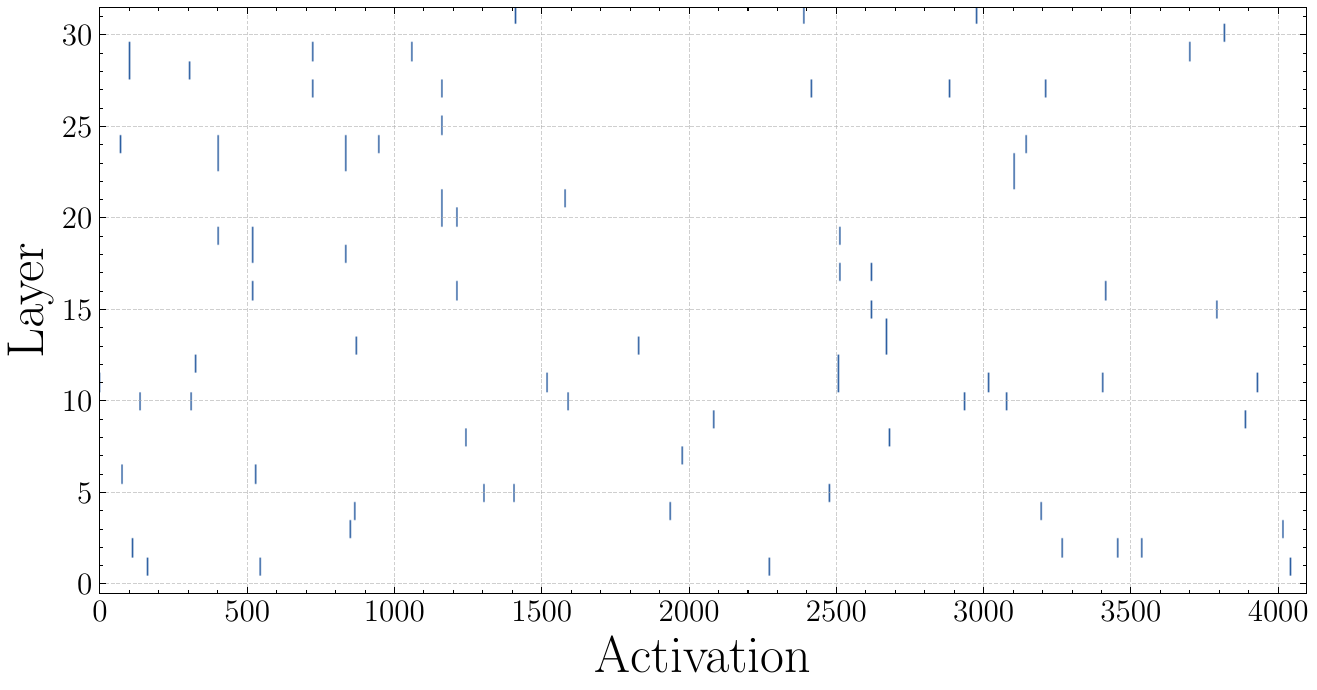}
         \caption{Overlapping \Ours{} between \textsc{InstaOrder (Depth.)} and \textsc{Clevr}.}
         \label{fig:ovl_instaorder_depth_clever}
     \end{subfigure}
     \hfill
     \begin{subfigure}[t]{0.49\textwidth}
         \centering
         \includegraphics[width=\textwidth]{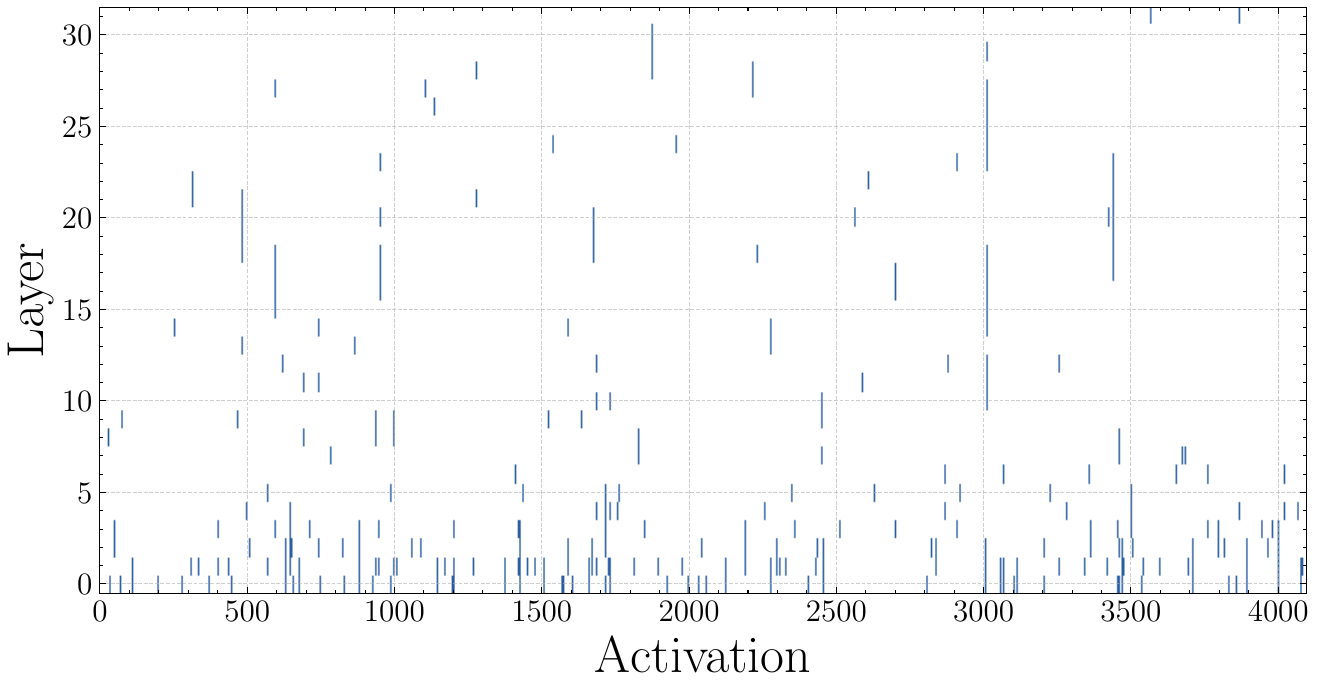}
         \caption{Overlapping \Ours{} between \textsc{InstaOrder (Occ.)}, \textsc{Vizwiz} and \textsc{Clevr}.}
         \label{fig:ovl_instaorder_occ_vizwiz_clevr}
     \end{subfigure}
     \hfill
     \hfill
     \begin{subfigure}[t]{0.49\textwidth}
         \centering
         \includegraphics[width=\textwidth]{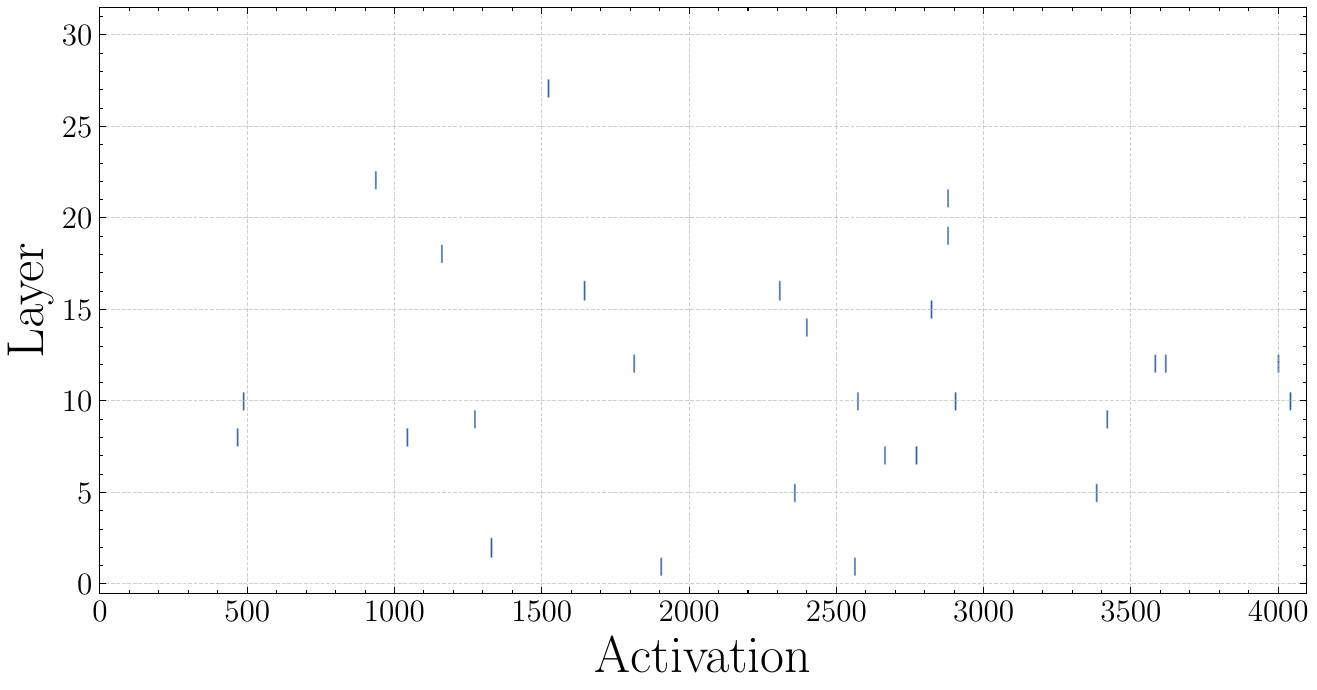}
         \caption{Overlapping \Ours{} between \textsc{InstaOrder (Occ.), \textsc{InstaOrder (Depth.)} and \textsc{Clevr}}.}
         \label{fig:ovl_instaorder_occ_instaorder_depth_clevr}
     \end{subfigure}
     \hfill
     \begin{subfigure}[t]{0.49\textwidth}
         \centering
         \includegraphics[width=\textwidth]{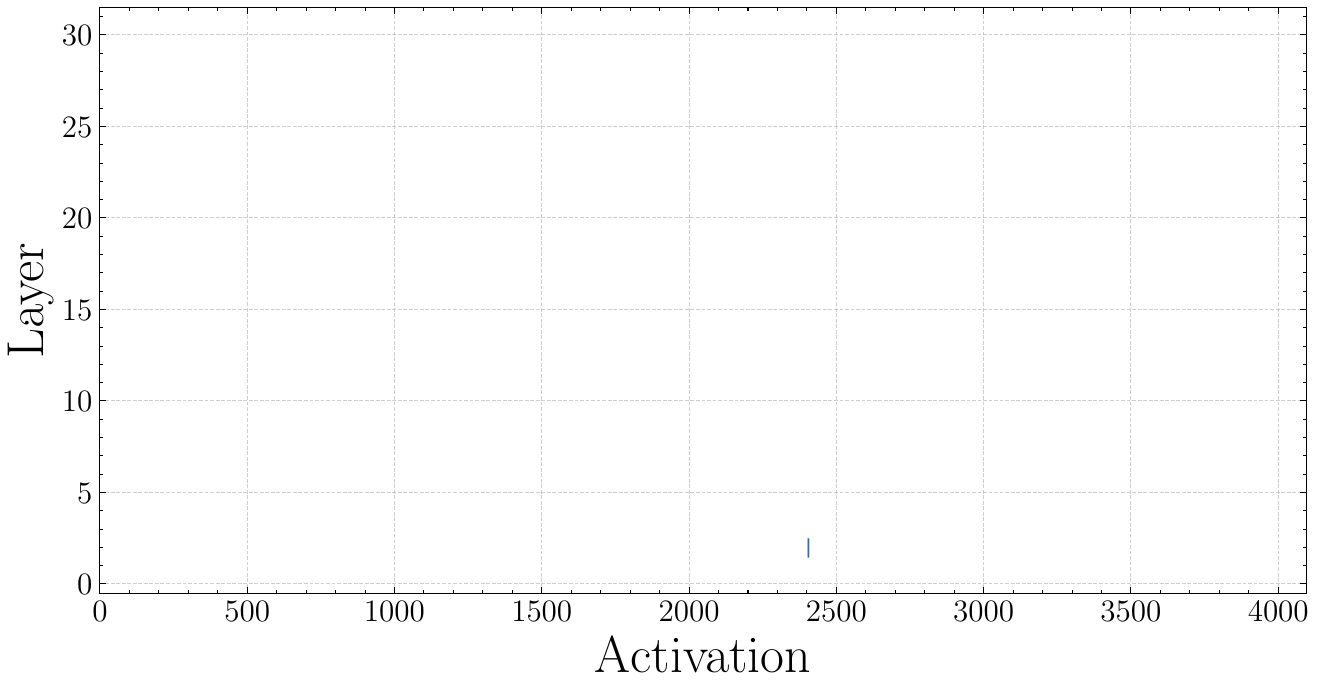}
         \caption{Overlapping \Ours{} between \textsc{InstaOrder (Occ.)}, \textsc{InstaOrder (Depth.)}, \textsc{Vizwiz} and \textsc{Clevr}.}
         \label{fig:ovl_instaorder_occ_instaorder_depth_vizwiz_clevr}
     \end{subfigure}
        \caption{\textbf{\Ours{} overlap between different datasets.} We visualize the probed neurons that surpass the model's performance on all the indicated datasets. We use \llavaSmall{}. \(x\)-axis indicate neuron index while \(y\)-axis indicates layer index. Depending on the dataset pair, many \Ours{} can overlap. We also find a quadruplet of datasets sharing \Ours{}.}
        \label{fig:overlap_sns}
\end{figure*}